%% file: main.tex
\def\isarxiv{1} 

\ifdefined\isarxiv
\documentclass[11pt]{article}

\usepackage[numbers]{natbib}

\else
\documentclass{article}
\usepackage{neurips_2023}

\fi

\usepackage{amsmath}
\usepackage{amsthm}
\usepackage{amssymb}
\usepackage{algorithm}
\usepackage{subfig}
\usepackage{algpseudocode}
\usepackage{graphicx}
\usepackage{grffile}
\usepackage{wrapfig,epsfig}
\usepackage{url}
\usepackage{xcolor}
\usepackage{epstopdf}

\usepackage{bbm}
\usepackage{dsfont}

\allowdisplaybreaks

\ifdefined\isarxiv

\usepackage{tikz}
\usepackage{hyperref}  
\hypersetup{colorlinks=true,citecolor=blue,linkcolor=blue} 
\usetikzlibrary{arrows}
\usepackage[margin=1in]{geometry}

\else

\usepackage{microtype}
\usepackage{hyperref}

\fi

\newtheorem{theorem}{Theorem}[section]
\newtheorem{lemma}[theorem]{Lemma}
\newtheorem{definition}[theorem]{Definition}

\newtheorem{corollary}[theorem]{Corollary}

\newtheorem{assumption}[theorem]{Assumption}

\newtheorem{fact}[theorem]{Fact}

\newtheorem{claim}[theorem]{Claim}

\newcommand{\wh}{\widehat}
\newcommand{\wt}{\widetilde}
\newcommand{\ov}{\overline}

\newcommand{\R}{\mathbb{R}}

\DeclareMathOperator{\poly}{poly}

\makeatletter
\newcommand*{\RN}[1]{\expandafter\@slowromancap\romannumeral #1@}
\makeatother

\usepackage{lineno}

\begin{document}

\ifdefined\isarxiv

\date{}

\title{Faster Robust Tensor Power Method for Arbitrary Order} 
\author{
Yichuan Deng\thanks{\texttt{ycdeng@cs.washington.edu}. The University of Washington.}
\and 
Zhao Song\thanks{\texttt{zsong@adobe.com}. Adobe Research.}
\and
Junze Yin\thanks{\texttt{junze@bu.edu}. Boston University.}
}

\else

\title{Faster Robust Tensor Power Method for Arbitrary Order} 
\maketitle 
\fi

\ifdefined\isarxiv
\begin{titlepage}
  \maketitle
  \begin{abstract}
\input{abstract}

  \end{abstract}
  \thispagestyle{empty}
\end{titlepage}

{
}
\newpage

\else

\begin{abstract}
\input{abstract}
\end{abstract}

\fi

\input{intro} 

\input{tech_ov}

\input{preli}

\input{robust_analysis}

\input{conclusion}

\ifdefined\isarxiv

\else
\bibliography{ref}
\bibliographystyle{plain}

\fi

\newpage
\onecolumn
\appendix

\section*{Appendix}

\input{app_preli}

\input{app_analysis}

\input{combine}

\input{sketch}

\ifdefined\isarxiv
\bibliographystyle{alpha}
\bibliography{ref}
\else

\fi




\end{document}

%% file: abstract.tex
Tensor decomposition is a fundamental method used in various areas to deal with high-dimensional data. \emph{Tensor power method} (TPM) is one of the widely-used techniques in the decomposition of tensors. This paper presents a novel tensor power method for decomposing arbitrary order tensors, which overcomes limitations of existing approaches that are often restricted to lower-order (less than $3$) tensors or require strong assumptions about the underlying data structure. We apply sketching method, and we are able to achieve the running time of $\widetilde{O}(n^{p-1})$, on the power $p$ and dimension $n$ tensor. We provide a detailed analysis for any $p$-th order tensor, which is never given in previous works.

%% file: intro.tex
\section{Introduction}
With the development of large-scale-data-driven applications, such as neural networks, social network analysis, and multi-media processing, 
\emph{tensors} have become a powerful paradigm to handle the data. According to \cite{swz16}, in recommendation systems, it's often beneficial to utilize more than two attributes to generate more accurate recommendations. For instance, in the case of Groupon, one could examine three attributes such as time, users, and  activities, which may include but are not limited to the factors like time of day, season, weekday, weekend, etc., as a basis for making predictions. More information on this can be found in \cite{kb09}.

Tensor decomposition is a mathematical tool that can break down the higher order tensor into a combination of lower order tensors. To deal with the high-dimensional data, decomposition becomes a natural method to handle the tensors, where the operation reads the original tensor as inputs and outputs the decomposition of it in some succinct form. 
Tensor decomposition has become a fundamental brick in many areas \cite{kb09}, including supervised and unsupervised learning \cite{aghkt14, jsa15}, reinforcement learning \cite{ala16}, statistics and computer vision \cite{sh05}, etc. More importantly, with the fast outbreak of Covid-19 and the rapid emergence of its new variants due to the large infectious population base, recent research applies the tensor model to analyze pandemic data \cite{dkn22} and use tensor decomposition to study the gene expression related to Covid-19 as in \cite{tt21}. Since gene expression is usually very complicated, tensor decomposition can efficiently help researchers to determine the connections between various variables to better understand complex systems, which may foster biological and medical development. It is beneficial to public health.

A well-known decomposition method is Candecomp/Parafac (CP) decomposition, also known as canonical polyadic decomposition \cite{h70, cc70}. In CP decomposition, the input tensor is decomposed into a set of rank $1$ components. Although decomposing arbitrary tensors is NP-hard \cite{hl13}, it can be made feasible for tensors with linearly independent components by using a whitening procedure to transform them into orthogonally decomposable tensors. The tensor power method (TPM) is a straightforward and effective technique for decomposing an orthogonal tensor and is an extension of the matrix power method. 

To be more specific, TPM requires calculating the inner product of two vectors, one of which is a rank-1 matrix and the other is a segment of a tensor. This type of inner product can be estimated much more efficiently because the sketch vectors have much lower dimensions and it is more convenient to compute the inner product of them. Additionally, it's possible to substitute sketching with sampling to approximate inner products \cite{swz16}. 

When there is no noise in the data, by performing random initialization followed by deflation, the tensor power method effectively can recover the components correctly. However, with the NP-hard nature of the decomposition of arbitrary tensors, the perturbation analysis of this method is more complex compared to the matrix case. When large amounts of arbitrary noise are added to an orthogonal tensor, the decomposition of the tensor will become intractable. Previous research has demonstrated guaranteed component recovery under bounded noise conditions in \cite{aghkt14}, with further improvements in \cite{agj17}. More recent work \cite{wa16} improved the noise requirement. 

Nowadays, there are two prevailing tensor decomposition research directions: 1). how to construct new machine learning related problems involving tensor decomposition, and 2). how to develop new tensor decomposition algorithms under weaker assumptions \cite{rsg17} or investigate fast tensor decomposition. Although tensor decomposition typically does not require strict conditions for uniqueness, their practical application in machine learning settings requires significant and stringent requirements as in \cite{wa16}.

\subsection{Our Result}

Therefore, to solve this issue for the purpose of better serving the important tensor decomposition applications as we mentioned before, we provide an algorithm that not only requires a milder assumption but also is suitable for wider tensor choices.

More specifically, we generalized the previous robust tensor power method for third-order tensors to general order tensors. That is, we provide an algorithm such that, for any \emph{arbitrary order} tensor $A \in \R^{n^p}$, it outputs the estimated eigenvector/eigenvalue pair, together with the deflated tensor. We present our main result as follows:

\begin{theorem}[Informal version]\label{thm:main_informal}
    There is a robust tensor power method (Algorithm~\ref{alg:importance_sampling_robust_power_method}) that takes any $p$-th order and dimension $n$ tensor as input, it uses $\wt{O}(n^p)$ spaces and $\wt{O}(n^p)$ time in initialization  and in each iteration, it only takes  $\wt{O}(n^{p-1})$ time.
\end{theorem}

\subsection{Related work}
\paragraph{Tensor decomposition.}

There is a series of works that focused on the efficient and fast decomposition of tensors \cite{t10, ptc13, cv14, hnh+13, kphf12, wlsh14, bs15}. Later work \cite{wtsa15} provided a method based on the random linear sketching technique to enable fast decomposition for orthogonal tensors. More recently, \cite{swz16} provided another approach of importance sampling, with a better running time. 

The canonical polyadic decomposition is a very famous and popular technique of decomposition. It is CANDECOMP/PARAFAC (CP) decomposition \cite{swz16}. In CP decomposition, a tensor can be broken down into a combination of rank-$1$ tensors that add up to it \cite{h70}, and this combination is the only possible one up to some minor variations, such as scaling and reordering of the tensors. In other words, there is only one way to decompose the tensor and any other arrangement of the rank-$1$ tensors that add up to the same tensor is not possible. This property of tensor decomposition is more restrictive than that of matrices, and it holds for a broader range of tensors. Therefore, tensor decomposition is considered to be more rigid than matrix decomposition. In \cite{wtsa15}, multiple applications, including computational neuroscience, data mining, and statistical learning, of tensor decomposition are mentioned. 

\paragraph{Unique tensor decomposition.}

Previous research in algebraic statistics has already linked tensor decompositions to the development of probabilistic models. By breaking down specific moment tensors using low rank decompositions, researchers could decide the extent of the identifiability of latent variable models \cite{amr09, aprs11, rs12}. The utilization of Kruskal's theorem in \cite{k77} was crucial in establishing the accuracy of identifying the model parameters. Nonetheless, this method supposes that people can use an infinite number of samples and cannot offer any information of what is the minimum sample size required to learn the model parameters in these given error bounds. Relying solely on Kruskal's theorem does not suffice to determine the bounds of sample complexity since by using it, we can only get that the low rank decompositions of actual moment tensors are unique, but we cannot get enough information about the decomposition of empirical moment tensors. Considering the necessary sample size to learn the parameters of the model, we need to get a uniqueness guarantee which is more robust. We need that this guarantee can satisfy that whenever $T'$, which is an empirical moment tensor, closely approximates $T$, which is a moment tensor, a low-rank decomposition of $T'$ would also closely resemble a low-rank decomposition of $T$.

\paragraph{Method of moments.}

Unique tensor decomposition in the method of moments is a very important application of matrix low rank approximation as shown in \cite{swz18, kmm22}. Tensor decomposition techniques suggest a different approach that relies on the method of moments, as described in \cite{rsg17}. The basic concept is to calculate the empirical moments of the data, such as the skewness, mean, and variance, and then determine the configuration of latent variables that would produce similar moments under the given model. Recent research has demonstrated that the low-order moment tensors of many common probabilistic models have a specific structure, as reported in \cite{agh+12}. This observation, combined with advances in multilinear algebra and tensors \cite{hh82, p99, ptc13, sl10, sbl13, tb06, bv18}, has allowed the development of effective and efficient algorithms for solving such problems. These methods are recognized for their ability to scale well to larger problems and generally do not suffer much from the curse of dimensionality, as discussed in \cite{ot09}.

\paragraph{Power method.}
The power method is a popular iterative algorithm for computing the dominant eigenvector and eigenvalue of a tensor. In recent years, there is a series of works \cite{l05, q05, q07, qsw07, cpz08, nqz10, wqz09} that focused on this topic. The work of \cite{km11} provides the result to compute real symmetric-tensor eigenpairs, which is closely related to the optimal rank-$1$ approximation of a symmetric tensor. Moreover, their method is based on the shifted symmetric higher-order power method (SS-HOPM), which can be viewed as a generalization of the power iteration method for matrices. \cite{aghkt14} considers the relation between tensor decomposition and learning latent variable models, where they also provided a detailed analysis of a robust tensor power method. More recent work by \cite{agj17} offers a new approach to analyzing the behavior of tensor power iterations in the overcomplete scenario, in which the tensor's CP rank surpasses the input dimension. 

\paragraph{Canonical/Polydic decomposition and Tucker decomposition.}

The most commonly employed techniques for breaking down tensors are CP (Canonical/Polydic) decomposition and Tucker factorization. CP decomposes a tensor that has higher order into a collection of fixed-rank individual tensors that are summed together, while Tucker factorization reduces a tensor that has higher order to a smaller core tensor and a matrix product of each of its modes. Non-negative tensor factorization is the extension of non-negative matrix factorization to multiple dimensions \cite{bks21}.

Recent research in Tucker decomposition has focused on developing more efficient algorithms for computing the decomposition \cite{pc08, zczx15, kc16}, improving its accuracy and robustness \cite{mc13, hcl22}, and applying it to various new domains, like image representation \cite{mc13}.

\paragraph{Sketching technique.}
Numerical linear algebra benefits greatly from sketching and sampling techniques, which have proven to be highly effective. These techniques have been successfully applied to a wide range of fundamental tasks. 
It also plays an important role in linear programming (LP) \cite{cls19,jswz21,y20,dly21,gs22}, tensor approximation \cite{swz19, mwz22, dgs23}, matrix completion
\cite{gsyz23}, matrix sensing \cite{qsz23,dls23}, submodular function maximization \cite{qsw23}, dynamic sparsifier \cite{djs+22}, dynamic tensor produce regression \cite{rsz22}, semi-definite programming \cite{syyz22_lichen}, sparsification problems involving an iterative process \cite{sxz22}, adversarial training \cite{gqls23}, kernel density estimation \cite{qrs+22}, distance oracle problem \cite{dswz22}, empirical risk minimization \cite{lsz19, qszz23} and also relational database \cite{qjs+22}. Recently, sketching technique also plays an important role in Large Language Models (LLMs) research \cite{dms23, dls23, gsy23, lsz23}.

\paragraph{Roadmap.}

In Section~\ref{sec:technique_overview}, we introduce the techniques used in this paper. In Section~\ref{sec:prelim}, we introduce the key concepts, notations, and properties which we use to support our analysis and new Theorems. In Section~\ref{sec:robust_tensor}, we present our main result. In Section~\ref{sec:conclusion}, we summarize this paper and provide some future research directions in this field. 

%% file: tech_ov.tex
\section{Technique Overview}
\label{sec:technique_overview}

In this particular section, we present a summary of the methods employed in our analysis. To be convenient, we will delay the definition of several math concepts but still use them here.

\paragraph{Loosened assumption.}
Our main breakthrough is that we generalize the robust tensor power method to support any order tensors. It efficiently resolves the drawback of the earlier method in \cite{wa16} that is limited in the tensor with the order below 3 and requires very strict assumptions. Moreover, we have created a strong and adaptable algorithm that can handle a variety of tensor data: natural language corpora, images, videos, etc. Then, we explain how we generalize this in detail. 

\paragraph{Recoverability of eigenvectors implied by bounded noise. }
Starting from the construction of the input tensor
\begin{align*}
    A  = A^* + \wt{E} \in \mathbb{R}^{n^p}
\end{align*}
where it consists of a part of decomposable tensor $A^*$ and a noise term $\wt{E}$, we show that, if the norm is bounded, in the form of 
\begin{align*}
    \| \wt{E}(I,u_t,\cdots,u_t)\|_2 \leq 6\epsilon /c_0
\end{align*}
and 
\begin{align*}
    |\wt{E}(v,u_t,\cdots,u_t)| \leq 6 \epsilon /(c_0 \sqrt{n}),
\end{align*}
then the compositions of $A^*$ is able to be recovered from $A$ (see details in Appendix~\ref{sec:combine}). Formally, the eigenvectors have the following properties: 
\begin{enumerate}
    \item The difference of the tangent from an eigenvector to the two unit vectors is bounded by a term $18\epsilon/ (c_0 \lambda_1)$ of the corresponding eigenvalue (see definition of $\tan \theta$ in Def.~\ref{def:sin_cos_tan}): 
    \begin{align*}
     \tan \theta(v_1, u_{t+1}) \leq 0.8 \tan \theta(v_1, u_{t}) + 18\epsilon/ (c_0 \lambda_1)
    \end{align*}
    \item Tail components are bounded by the top component, in the power of $p-2$: 
    \begin{align*}
        \underset{j\in [k]\backslash\{1\}}{\max} \lambda_j |v_j^\top u_t |^{p-2} \leq (1/4) \lambda_1 |v_1^\top u_t|^{p-2}.
    \end{align*}
    \item With all $j$ being an arbitrary element in $\{2,\cdots,k\}$, 
    \begin{align*}
        |v_j^\top u_{t+1} | / | v_1^\top u_{t+1} | \leq ~  0.8 |v_j^\top u_t|  / |v_1^\top u_t|  + 18 \epsilon /(c_0 \lambda_1 \sqrt{n})
    \end{align*}
\end{enumerate}
As these are generalized statements from previous results \cite{wa16, aghkt14} from bounded order ($p \le 3$) to general order $p$, the proof requires a much different analysis. We described the detail of our approach in the following paragraph.

\paragraph{Analysis of the recoverability. }

To show part 1 (see the details in Appendix~\ref{sec:more_analysis}), we have to upper bound $\tan \theta(v_1, u_{t+1})$. We first turn the tangent into terms of sine and cosine, which can be represented by the norm of the tensors. Then by simply using Cauchy-Schwartz, we can upper bounded the term by
\begin{align*}
      \tan\theta(v_1, u_{t+1}) 
     \le  \frac{ \| V^\top A ^*(I,u_t,\cdots, u_t) \|_2 + \| V^\top \wt{E}_{u_t} \|_2 }{ |v_1^\top A ^*(I, u_t, \cdots, u_t ) | - |v_1^\top \wt{E}_{u_t}| }. 
\end{align*}
Using a property for orthogonal tensor that, for $A ^*=\sum_{j=1}^k \lambda_j v_j^{\otimes p} \in \mathbb{R}^{n^p}$, it holds that for any $j \in [k]$,
\begin{align*}
    |v_j^\top A ^*(I,u,\cdots,u)| = \lambda_j |v_j^\top u|^{p-1}, 
\end{align*}
we are able to upper bound $\tan \theta(v_1, u_{t+1})$ with $\tan \theta(v_1, u_{t})$ in the form of 
\begin{align*}
    \tan \theta(v_1, u_{t+1}) \le \tan \theta(v_1, u_{t}) \cdot \frac{1}{4} \cdot B_1 + B_1 \cdot B_2,
\end{align*}
where $B_1$ and $B_2$ are two simplified terms defined as
\begin{align*}
    B_1 := & ~ \frac{1}{1-|v_1^\top \wt{E}_{u_t}| /(\lambda_1 |v_1^\top u_t|^{p-1}) } \\
    \text{and~}B_2 := & ~ \frac{\| V^\top \wt{E}_{u_t}\|_2 }{\lambda_1 |v_1^\top u_t|^{p-1}}. 
\end{align*}
By careful nontrivial analysis, we further show that these two terms can be bounded by 
\begin{align*}
    B_1 \le & ~ 1.1 \\
    B_2 \le & ~ 18 \epsilon/(c_0\lambda_1).
\end{align*}
Combining all these, we are able to complete the proof of the first property.

The second part is relatively simple than the first part. Using the same property for orthogonal tensor, we start by lower bounding the term 
\begin{align*}
    \frac{|v_1^\top u_{t+1}|}{v_j^\top u_{t+1}} 
    \ge \frac{ \frac{9}{10} |v_1^\top u_t|}{ \frac{1}{4} |v_j^\top u_t| + \frac{1}{10} |v_1^\top u_t|  }.
\end{align*}
We then divide the proof into two conditions. 
First, if $|v_j^\top u_t| < |v_1^\top u_t|$, then the proportion of the top component over other rest components can be easily lower bounded by, 
\begin{align*}
    \frac{\lambda_1|v_1^\top u_{t+1}|^{p-2}}{\lambda_j|v_j^\top u_{t+1}|^{p-2}} \ge \frac{\lambda_1}{\lambda_j}\cdot 2^{p-2}. 
\end{align*}
For the opposite condition that $|v_j^\top u_t| \ge |v_1^\top u_t|$, we give a better analysis than previous work (see \cite{wa16}'s Lemma~C.2). We show that for all $p$ being greater than or equal to $3$, it holds that
\begin{align*}
    \frac{\lambda_1|v_1^\top u_{t+1}|^{p-2}}{\lambda_j|v_j^\top u_{t+1}|^{p-2}} \ge 4\cdot 2^{p-2}. 
\end{align*}

The final property is also proved in a similar way. For simplicity, we first define two terms
\begin{align*}
    B_3 := & ~ \frac{1}{1 - |v_1^\top \wt{E}_{u_t}|/(\lambda_1|v_1^\top u_t|^{p-1})} \\
    B_4 := & ~ \frac{|v_j^\top \wt{E}_{u_t}|}{\lambda_1|v_1^\top u_t|^{p-1}}.
\end{align*}
Similarly, as in the analysis before, we upper bound 
\begin{align*}
    \frac{|v_j^\top u_{t+1}|}{|v_1^\top u_{t+1}|} \le \frac{|v_j^\top u_{t}|}{|v_1^\top u_{t}|} \cdot \frac{1}{4} B_3 + B_3 \cdot B_4. 
\end{align*}
$B_3$ can be easily bounded by a similar proof if $|v_1^\top u_t| \ge 1 - \frac{1}{c_0^2p^2k^2}$. For $B_4$, we divide it into two case: $|v_1^\top u_t| \le 1 - \frac{1}{c_0^2p^2k^2}$. By a different discussion, we can show that $B_4 \le 18\epsilon/(c_0\lambda_1\sqrt{n})$. 

\paragraph{Bounding the recovery error.}
We now step to the final technical lemma which shows the bound of the approximation error of the output of our algorithm: 

\begin{lemma}[Informal version]\label{lem:informal_version}
If the following conditions hold:
\begin{itemize}
    \item Let $p$ be greater than or equal to $3$.
    \item Let $k$ be greater than or equal to $1$.
    \item Let $A  = A ^* +E \in \mathbb{R}^{n^p}$ be an arbitrary tensor.
    \item Suppose that $\lambda_{1}$ is the greatest values in $\{\lambda_i\}_{i=1}^k$.
    \item Suppose that $\lambda_{k}$ is the smallest values in $\{\lambda_i\}_{i=1}^k$.
    \item The outputs obtained from the robust tensor power method are $\{\wh{\lambda}_i, \wh{v}_i\}_{i=1}^k$.
    \item $E$ satisfies that $\| E \| \leq \epsilon / (c_0 \sqrt{n} )$ (see definition of $\| E \|$ in Section~\ref{sec:prelim}).
    \item $T = \Omega( \log(\lambda_{1}  n/\epsilon) )$.
    \item $L = \Omega( k \log(k))$.
\end{itemize}
  Then, there exists a permutation $\pi:[k] \rightarrow [k]$, such that $\forall i\in[k]$,
  \begin{align*}
  |\lambda_i - \wh{\lambda}_{\pi(i)}| \leq \epsilon, \ \ \ \
  \|v_i - \wh{v}_{\pi(i)}\|_2 \leq \epsilon/ \lambda_i.
  \end{align*}

 \end{lemma}

This Lemma is the key component of our main Theorem (Theorem~\ref{thm:main_informal}). We use mathematical induction to show this Lemma. To show the base case, we need to bound three different terms, namely $|\wh{v}_1 - v_1|$, $|\wh{\lambda}_1 - \lambda_1|$, and $|\wh{v}_1^\top v_j|$.

To bound $|\wh{v}_1 - v_1|$, we need to utilize the properties of angle and apply the definitions and Lemmas we develop in Section~\ref{sec:robust_tensor}. First, we can show
\begin{align*}
    \tan \theta(u_0, v_1) \leq \sqrt{n}.
\end{align*}
By using the fact that $|u_{t^*}^\top v_1 | = 1 - \frac{1}{c_0^2 p^2 k^2}$ together with some respective properties of $u_{t^*}^\top$ and $v_1$, we can get
\begin{align*}
    \| u_{t^*}- v_1 \|_2^2 = 2/(c_0^2 p^2 k^2).
\end{align*}
Finally, we can bound $|\wh{v}_1 - v_1|$ by using the these information and recursively applying Part 1 of Theorem~\ref{thm:general_p_bound_EIu}.

For the second term $|\wh{\lambda}_1 - \lambda_1|$, we try to simplify it and split that into three parts, namely $B_5$, $B_6$, and $B_7$ which are defined as follows
\begin{align*}
    B_5 &:= | \wt{E}(\wh{v}_1,\cdots, \wh{v}_1)|\\
    B_6 &:= |\lambda_1 |v_1^\top \wh{v}_1|^p -\lambda_1|\\
    B_7 &:= \sum_{j=2}^k \lambda_j |v_j^\top \wh{v}_1|^p.
\end{align*}
It suffices to bound these three terms, but bounding these is challenging. We prove that $B_5 \leq \epsilon/12$, $B_6 \leq \epsilon/12$, and $B_7 \leq \epsilon/4$. By putting these together, we get that 
\begin{align*}
    |\wh{\lambda}_1 - \lambda_1| \leq \epsilon/12 + \epsilon/12 + \epsilon/4 \leq \epsilon.
\end{align*}
Moreover, we need to give $\epsilon$ a proper value. If $\epsilon$ is too big, we might not get our desired result. On the other hand, if $\epsilon$ is too small, the result might be meaningless. Finally, by setting $\epsilon < \frac{1}{4} k^{1/(p-1)} \lambda_k$, we get the desired result.

What is left out is the third term $|\wh{v}_1^\top v_j|$. We need to recursively apply the third part of Theorem~\ref{thm:general_p_bound_EIu}. We show that
\begin{align*}
 |v_j^\top u_{t^*}| / |v_1^\top u_{t^*}| \leq 0.8^{t^*} \cdot 1 / (1/\sqrt{n}).
\end{align*}
In the end, by choosing proper $T$ and $t^*$ values, we can get our desired bound.

In the inductive case, the arrangement of the proof is just like the ones in the base case: we also need to bound these three terms. Moreover, for $i$ being larger, we also need to consider the noise, namely
\begin{align*}
    \wt{E} = E+ \sum_{i=1}^r E_i + \ov{E}\in \mathbb{R}^{n^p},
\end{align*}
which adds more complexity to the condition we encounter. 

\paragraph{Sketching technique.}
Inspired by a recent sketching technology \cite{cn22}, we apply the similar sketching operation to develop a distance estimation data structure to apply in our tensor power method. Our data structure uses Randomized Hadamard Transform (RHT) to generate the sketching matrix. The data structure stores the sketches of a set of maintained tensors $\{A_i\}_{i \in [n]} \subseteq \R^{n^{p-1}}$. Suppose we have already known the decomposition of a tensor $A_i$, i.e.,
\begin{align*}
    \sum_{j=1}^k \alpha_{i,j} x_j^{\otimes (p-1)}.
\end{align*}
Now when a query tensor in the form of
$
    q = u^{\otimes (p-1)}
$ 
comes, our data structure can read 
\begin{align*}
    \{x_j\}_{j \in [k]}, \alpha\in\R^{n \times k}, u \in \R^n,
\end{align*}
and return an $(1 \pm \epsilon)$ estimated product $v \in \R^n$ such it approximates
\begin{align*}
    \langle A_i - \sum_{j=1}^k \alpha_{i,j} x_j^{\otimes (p-1)} , u^{\otimes (p-1)} \rangle.
\end{align*}
This procedure runs fast in time $\wt{O}(\epsilon^{-2}n^{p-1} + n^2k)$. Applying this data structure when computing the error, we are able to achieve our final fast TPM algorithm. 

%% file: preli.tex
\section{Preliminary}
\label{sec:prelim}

In this section, we first introduce the notations we use. Then, we elucidate the fundamental concepts. $\R$ denotes the set that contains all real numbers. 
For a scalar $a$, i.e. $a \in \R$, $|a|$ represents the absolute value of $a$.
For any $A \in \R^{n \times k}$ being a matrix and $x \in \R^k$ being a vector, we use $\| A \|:= \max_{x \in \R^k} \| A x \|_2 / \| x \|_2$ to denote the spectral norm of $A$. 
We use $\| x \|_2:= (\sum_{i=1}^n x_i^2)^{1/2}$ to denote the $\ell_2$ norm of vector $x$.
For two vectors $u \in \R^n$ and $v \in \R^n$, we use $\langle u, v \rangle$ to denote inner product, i.e., $\langle u , v \rangle = \sum_{i=1}^n u_i v_i$.
Let $p \geq 1$ denote some integer. We say $E \in \R^{n \times \cdots \times n}$ (where there are $p$ of $n$), if $E$ is a $p$-th order tensor and every dimension is $n$. For simplicity, we write $E \in \R^{n^p}$. If $p=1$, then $E$ is a vector. If $p=2$, then $E \in \R^{n \times n}$ is a matrix. If $p=3$, then $E \in \R^{n \times n \times n}$ is a 3rd order tensor.
For any two vectors $x,y$, we define $\theta(x,y)$ to be that $\cos \theta(x,y) = \langle x,y\rangle$.
For a 3rd tensor $E \in \R^{n \times n \times n}$, we have $E(a,b,c) \in \R$
\begin{align*}
E(a,b,c)= \sum_{i=1}^n \sum_{j=1}^n \sum_{k=1}^n E_{i,j,k} a_{i} b_{j} c_{k}.
\end{align*}
We define $\| E \|:= \max_{x : \| x \|_2=1} |E(x,x,x)|$. 
Similarly, the definition can be generalized to $p$-th order tensor.
For a 3rd order tensor $E \in \R^{n \times n \times n}$, we have $E(I,b,c) \in \R^n$,
\begin{align*}
E(I,b,c)_i =  \sum_{j=1}^n \sum_{k=1}^n E_{i,j,k} b_j c_k, ~~~\forall i \in [n]
\end{align*}
For a 3rd order tensor $E \in \R^{n \times n \times n}$, we have $E(I,I,c)\in \R^{n \times n}$
\begin{align*}
E(I,I,c)_{i,j} = \sum_{k=1}^n E_{i,j,k} c_k, ~~~ \forall i, j \in [n] \times [n].
\end{align*} 
Let $a,b,c\in \R^n$. Let $E = a \otimes b \otimes c \in \R^{n \times n \times n}$. We have
\begin{align*}
E_{i,j,k} = a_i b_j c_k , ~~~\forall i \in [n], \forall j \in [n], k \in [n]
\end{align*}
Let $a \in \R^n$, let $E = a\otimes a \otimes a = a^{\otimes 3} \in \R^{n \times n \times n}$. We have
\begin{align*}
E_{i,j,k}  = a_i a_j a_k,~~~\forall i \in [n], \forall j \in [n], k \in [n]
\end{align*}
Let $E = \sum_{i=1}^m u_i^{\otimes 3}$. Then we have $E(a,b,c) \in \R$
\begin{align*}
E (a,b,c) = \sum_{i=1}^m ( u_i^{\otimes 3} (a,b,c) ) = \sum_{i=1}^m \langle u_i, a\rangle \langle u_i, b \rangle \langle u_i, c \rangle.
\end{align*}
For $\mu \in \R^d$ and $\Sigma \in \R^{n \times n}$. We use ${\cal N}(\mu, \Sigma)$ to denote a Gaussian distribution with mean $\mu$ and covariance $\Sigma$. For $x \sim {\cal N}(\mu, \Sigma)$, we way it is a Gaussian vector.
For any vector $a \in \R^n$, we use $\max_{i \in [n]} a_i$ to denote a value $b$ over sets $\{a_1, a_2, \cdots, a_n\}$.
For any vector $a \in \R^n$, we use $\arg\max_{i\in [n]} a_i$ to denote the index $j$ such that $a_j = \max_{i \in [n]} a_i$.

\begin{fact}\label{fac:x_p}
 We have
\begin{itemize}
    \item Part 1. For any $x \in (0,1)$ and integer $p \geq 1$, we have $| 1 - (1-x)^p | \leq p \cdot x$.
    \item Part 2. $(a+b)^p \leq 2^{p-1} a^p + 2^{p-1} b^p$.
\end{itemize}
 \end{fact}

%% file: robust_analysis.tex
\section{Robust Tensor Power Method Analysis for General Order \texorpdfstring{$p\geq 3$}{}}

\label{sec:robust_tensor}

\begin{algorithm}[!ht]
\caption{Our main algorithm}\label{alg:importance_sampling_robust_power_method}
\begin{algorithmic}[1]
\Procedure{FastTensor}{$A$}
\State $\mathrm{ds}.\textsc{Init}(A)$
\For{$\ell = 1 \to L$}
	\For{$t=1 \to T$}
		\State $u^{(\ell)} \gets \mathrm{ds}.\textsc{Query}(u^{(\ell)})$ \Comment{Lemma~\ref{lem:data_structure}}
		\State $u^{(\ell)} \leftarrow u^{(\ell)} / \| u^{(\ell)} \|_2$
	\EndFor
	\State $\lambda^{(\ell)} \leftarrow  \mathrm{ds}.\textsc{QueryValue}(u^{(\ell)})$ \Comment{Lemma~\ref{lem:data_structure}}
\EndFor
\State $\ell^* \leftarrow \arg\max_{\ell \in [L]} {\lambda^{(\ell)}}$
\State $u^* \leftarrow u^{(\ell^*)}$
\For{$t=1 \to T$}
	\State $u^* \leftarrow \mathrm{ds}.\textsc{Query}( u^* )$
	\State $u^* \leftarrow u^* / \| u^* \|_2$
\EndFor
\State $\lambda^* \gets \mathrm{ds}.\textsc{QueryValue}(u^*)$

\State {\bf return}{ $\lambda^*, u^*$}
\EndProcedure
\end{algorithmic}
\end{algorithm}

In Section~\ref{sec:general_p_useful_facts}, we analyze the properties of the $p$-th order tensor, where $p$ is an arbitrary positive integer greater than or equal to $3$. These properties are generalized from the third and the fourth order tensors. In Section~\ref{sec:general_p_key_lemmas}, we generalize the properties of the existing robust tensor power method from the third order to any arbitrary order greater than or equal to three.

\begin{definition}\label{def:sin_cos_tan}
For $u,v$ be unit vectors, we define 
\begin{itemize}
    \item $\cos\theta(u,v) := \langle u, v \rangle$ 
    \item $\sin \theta(u,v) := \sqrt{1-\cos^2 \theta(u,v)}$
    \item $\tan \theta(u,v) := \sin \theta(u,v) / \cos\theta(u,v)$
\end{itemize}
\end{definition}

\subsection{Useful facts}\label{sec:general_p_useful_facts}

In this section, we introduce important facts.

\begin{fact}[Informal version of Fact~\ref{fac:vj_dot_TIu:formal}]\label{fac:vj_dot_TIu:informal}
If the following conditions hold
\begin{itemize}
    \item Let $p$ be greater than or equal to $3$.
    \item Let $A ^*=\sum_{j=1}^k \lambda_j v_j^{\otimes p} \in \mathbb{R}^{n^p}$ be the orthogonal tensor.
    \item Given a vector $u\in \mathbb{R}^n$.
    \item Let $j\in [k]$.
\end{itemize}

Then, we can get 
\begin{align*}
    |v_j^\top A ^*(I,u,\cdots,u)| = \lambda_j |v_j^\top u|^{p-1}.
\end{align*}

\end{fact}

\begin{fact}[Informal version of Fact~\ref{fac:bounding_E_p:formal}]\label{fac:bounding_E_p:informal}
If the following conditions hold
\begin{itemize}
    \item $E \in \mathbb{R}^{n^p}$ is an arbitrary tensor.
    \item $u,v\in \mathbb{R}^n$ are two arbitrary unit vectors.
\end{itemize}

Then,
\begin{enumerate}
\item $| E(u,v,\cdots,v) | \leq \| E \|$.
\item $\| E(I,v,\dots,v) \|_2 \leq \sqrt{n} \| E\|$.
\end{enumerate}
\end{fact}

\begin{fact}[Informal version of Fact~\ref{fac:diff_two_tensor_l2:formal}]\label{fac:diff_two_tensor_l2:informal}
If the following conditions hold
\begin{itemize}
    \item $p$ is greater than or equal to $3$.
    \item $x,y,u,v\in \mathbb{R}^n$ are arbitrary unit vectors.
    \item Let $j\in \{0,1,\cdots,p-2\}$.
\end{itemize}

Then, we have
\begin{align}\label{eq:equal_x-y}
\| [x\otimes v^{\otimes (p-1)}](I,u,\cdots,u) -  [y\otimes v^{\otimes (p-1)}](I,u,\cdots,u) \|_2 
=  |\langle u, v\rangle|^{p-1} \cdot \| x - y\|_2
\end{align}
and
\begin{align}\label{eq:leq_x-y}
 & ~ \| [  v^{\otimes (1+j)} \otimes x \otimes v^{\otimes (p-2-j)}](I,u,\cdots,u)  - [v^{\otimes (1+j)} \otimes y \otimes v^{\otimes (p-2-j)}](I,u,\cdots,u) \|_2 \notag\\
\leq & ~ |\langle u,v \rangle|^{p-2} \cdot \| x - y \|_2.
\end{align} 
\end{fact}

\begin{fact}[Informal version of Fact~\ref{fac:V_dot_T:formal}]\label{fac:V_dot_T:informal}

If the following conditions hold
\begin{itemize}
    \item $v_1, v_2, \cdots, v_n$ is an orthonormal basis.
    \item $V=(v_2,\cdots,v_n) \in \mathbb{R}^{n\times (n-1)}$.
    \item $A ^* = \sum_{i=1}^k \lambda_i v_i^{\otimes p}$.
    \item Let $u\in \mathbb{R}^n$ be a vector.
\end{itemize}

Then, we have 
\begin{align*}
    \| V^\top A ^*(I,u,\cdots, u) \|_2^2 = \sum_{j=2}^k \lambda_j^2 |v_j^\top u|^{2(p-1)}.
\end{align*}
\end{fact}

\subsection{Convergence guarantee and deflation}\label{sec:general_p_key_lemmas}

In this section, we present our main result: we generalize the robust tensor power method to all the situations where $p$ is greater than or equal to $3$.

\begin{lemma}
\label{lem:wang_lem_C1}
If the following conditions hold:
\begin{itemize}
    \item Let $t\in [k]$.
    \item Let $\eta \in (0,1/2)$.
    \item In $\mathbb{R}^n$, ${\cal U}$ represents a set of random Gaussian vectors. 
    \item $|{\cal U}| = \Omega(k \log(1/\eta))$.
\end{itemize}

Then, there is a probability of at least $1-\eta$ that there exists a vector $u \in {\cal U}$ satisfying the following condition:
\begin{align*}
 \underset{j\in [k] \backslash \{t\}}{\max} | v_j^\top u| \leq \frac{1}{4} |v_t^\top u|  \mathrm{~and~}  |v_t^\top u | \geq 1/\sqrt{n}.
\end{align*}
\end{lemma}

We analyze \cite{wa16}'s Lemma C.2 and generalize it from $p$ being equal to $3$ to any $p$ being greater than or equal to $3$.

In the following Theorem, intuitively, we treat $A^*$ as the ground-truth tensor. We treat $\wt{E}$ as the noise tensor. In reality, we can not access the $A^*$ directly. We can only access $A^*$ with some noise which is $\wt{E}$. But whenever $\wt{E}$ (the noise) is small compared to ground-truth $A^*$, then we should be able to recover $A^*$.  

\begin{theorem}\label{thm:general_p_bound_EIu}
If the following conditions hold
\begin{itemize}
    \item Let $\wt{E} \in \R^{n^p}$ denote some tensor. 
    \item $c > 0$ is an arbitrarily small number.
    \item Let $c_0\geq 1$.
    \item Let $p$ be greater than or equal to $3$.
    \item $A  =A ^*+\wt{E} \in \mathbb{R}^{n^p}$ is an arbitrary tensor satisfying $A ^*=\sum_{i=1}^k\lambda_i v_i^{\otimes p}$.
    \item Let $u_{t+1} = \frac{A (I,u_{t},\cdots,u_{t})}{\| A (I,u_{t},\cdots,u_{t}) \|_2}$.
    \item We define Event $\xi$ to be $|v_1^\top u_t| \leq 1- 1/(c_0^2 p^2 k^2)$.
    \item Let $\epsilon \leq \frac{c \lambda_1}{(c_0 p^2 k n^{(p-2)/2})}$.
    \item $u_t\in \mathbb{R}^n$ is an unit vector
    \item Let $t\in [T]$.
    \item Suppose \begin{align*}
\| \wt{E}(I,u_t,\cdots,u_t)\|_2 \leq 
\begin{cases}
    4p\epsilon ,     &  \mathrm{if~} \xi \\
    6\epsilon /c_0 ,  &  \mathrm{ow}. \\
  \end{cases}
\end{align*}
and
\begin{align*}
|\wt{E}(v,u_t,\cdots,u_t)| \leq 
\begin{cases}
    4 \epsilon /\sqrt{n} & \quad \mathrm{if~} \xi \\
    6 \epsilon /(c_0 \sqrt{n})  & \quad \mathrm{ow}\\
\end{cases}
\end{align*}
\end{itemize}

Then,
\begin{enumerate}
\item 
\begin{align}\label{eq:part_1_of_thm}
  ~ \tan \theta(v_1, u_{t+1}) 
 \leq & ~
  \begin{cases}
    0.8 \tan \theta(v_1, u_{t})       & \quad \mathrm{if~} \xi \\
    0.8 \tan \theta(v_1, u_{t}) + 18\epsilon/ (c_0 \lambda_1) & \quad \mathrm{ow} \\
  \end{cases}
\end{align}
\item 
\begin{align}\label{eq:part_2_of_thm}
    \underset{j\in [k]\backslash\{1\}}{\max} \lambda_j |v_j^\top u_t |^{p-2} \leq (1/4) \lambda_1 |v_1^\top u_t|^{p-2}.
\end{align}
\item For any $j\in \{2,\cdots,k\}$,
\begin{align}\label{eq:part_3_of_thm}
   |v_j^\top u_{t+1} | / | v_1^\top u_{t+1} | 
 \leq 
  \begin{cases}
    0.8 |v_j^\top u_t| / |v_1^\top u_t|        & \quad \mathrm{if~} \xi \\
    0.8 |v_j^\top u_t| / |v_1^\top u_t|  + 18 \epsilon /(c_0 \lambda_1 \sqrt{n}) & \quad \mathrm{ow}
  \end{cases}
\end{align}
\end{enumerate}
\end{theorem}

%% file: conclusion.tex
\section{Conclusion}
\label{sec:conclusion}
We present a robust tensor power method that supports arbitrary order tensors. Our method overcomes the limitations of existing approaches, which are often restricted to lower-order tensors or require strong assumptions about the underlying data structure. By leveraging advanced techniques from optimization and linear algebra, we have developed a powerful and flexible algorithm that can handle a wide range of tensor data, from images and videos to multivariate time series and natural language corpora. We believe that our result has some insights on various tasks, including tensor decomposition, low-rank tensor approximation, and independent component analysis. We believe that our contribution will significantly advance the field of tensor analysis and provide new opportunities for handling high-dimensional data in various domains. We here propose some future directions. We encourage extending our method to more challenging scenarios, such as noisy data analysis, and exploring its applications in emerging areas, such as neural networks and machine learning. Finally, our results are totally theoretical. We are not aware of any negative societal impact.

%% file: app_preli.tex
\paragraph{Roadmap.}

In Section~\ref{sec:appen_prelim}, we introduce the background concepts (definitions and properties) that we use in the Appendix. In Section~\ref{sec:more_analysis}, we provide more details and explanations to support the properties we developed in this paper. In Section~\ref{sec:combine}, we present our important Theorems (Theorem~\ref{thm:general_p_without_sketch} and Theorem~\ref{thm:main_formal}) and their proofs. In Section~\ref{sec:fast_sketching_data_structure}, we provide the fast sketching data structure.

\section{Preliminary}
\label{sec:appen_prelim}

In Section~\ref{sec:appen_prelim:notation}, we define several basic notations.
In Section~\ref{sec:appen_prelim:fact}, we state several basic facts. In Section~\ref{sec:appen_prelim:tensor}, we present tensor facts and tools.

\subsection{Notations}\label{sec:appen_prelim:notation}

In this section, we start to introduce the fundamental concepts we use. 

For any function $f$, we use $\wt{O}(f)$ to denote $f \cdot \poly (\log f)$.

$\R$ denotes the set that contains all real numbers. 

For a scalar $a$, i.e. $a \in \R$, $|a|$ represents the absolute value of $a$.

For any $A \in \R^{n \times k}$ being a matrix and $x \in \R^k$ being a vector, we use $\| A \|:= \max_{x \in \R^k} \| A x \|_2 / \| x \|_2$ to denote the spectral norm of $A$. 

We use $\| x \|_2:= (\sum_{i=1}^n x_i^2)^{1/2}$ to denote the $\ell_2$ norm of vector $x$.

For two vectors $u \in \R^n$ and $v \in \R^n$, we use $\langle u, v \rangle$ to denote inner product, i.e., $\langle u , v \rangle = \sum_{i=1}^n u_i v_i$.

Let $p \geq 1$ denote some integer. We say $E \in \R^{n \times \cdots \times n}$ (where there are $p$ of $n$), if $E$ is a $p$-th order tensor and every dimension is $n$. For simplicity, we write $E \in \R^{n^p}$. If $p=1$, then $E$ is a vector. If $p=2$, $E \in \R^{n \times n}$ is a matrix. If $p=3$, then $E \in \R^{n \times n \times n}$ is a 3rd order tensor.

For any two vectors $x,y$, we define $\theta(x,y)$ to be that $\cos \theta(x,y) = \langle x,y\rangle$.

For a 3rd tensor $E \in \R^{n \times n \times n}$, we have $E(a,b,c) \in \R$
\begin{align*}
E(a,b,c)= \sum_{i=1}^n \sum_{j=1}^n \sum_{k=1}^n E_{i,j,k} a_{i} b_{j} c_{k}.
\end{align*}
Similarly, the definition can be generalized to $p$-th order tensor.

For a 3rd order tensor $E \in \R^{n \times n \times n}$, we have $E(I,b,c) \in \R^n$,
\begin{align*}
E(I,b,c)_i =  \sum_{j=1}^n \sum_{k=1}^n E_{i,j,k} b_j c_k, ~~~\forall i \in [n]
\end{align*}

For a 3rd order tensor $E \in \R^{n \times n \times n}$, we have $E(I,I,c)\in \R^{n \times n}$
\begin{align*}
E(I,I,c)_{i,j} = \sum_{k=1}^n E_{i,j,k} c_k, ~~~ \forall i, j \in [n] \times [n].
\end{align*}

Let $a,b,c\in \R^n$. Let $E = a \otimes b \otimes c \in \R^{n \times n \times n}$. We have
\begin{align*}
E_{i,j,k} = a_i b_j c_k , ~~~\forall i \in [n], \forall j \in [n], k \in [n]
\end{align*}

Let $a \in \R^n$, let $E = a\otimes a \otimes a = a^{\otimes 3} \in \R^{n \times n \times n}$. We have
\begin{align*}
E_{i,j,k}  = a_i a_j a_k,~~~\forall i \in [n], \forall j \in [n], k \in [n]
\end{align*}

Let $E = \sum_{i=1}^m u_i^{\otimes 3}$. Then we have $E(a,b,c) \in \R$
\begin{align*}
E (a,b,c) = \sum_{i=1}^m ( u_i^{\otimes 3} (a,b,c) ) = \sum_{i=1}^m \langle u_i, a\rangle \langle u_i, b \rangle \langle u_i, c \rangle.
\end{align*}

For $\mu \in \R^d$ and $\Sigma \in \R^{n \times n}$. We use ${\cal N}(\mu, \Sigma)$ to denote a Gaussian distribution with mean $\mu$ and covariance $\Sigma$. For $x \sim {\cal N}(\mu, \Sigma)$, we way it is a Gaussian vector.

For any vector $a \in \R^n$, we use $\max_{i \in [n]} a_i$ to denote a value $b$ over sets $\{a_1, a_2, \cdots, a_n\}$.

For any vector $a \in \R^n$, we use $\arg\max_{i\in [n]} a_i$ to denote the index $j$ such that $a_j = \max_{i \in [n]} a_i$.

Let $\mathbb{N}$ denote non-negative integers. 

\subsection{Basic Facts}\label{sec:appen_prelim:fact}

In this section, we introduce some basic facts.

\begin{fact}[Geometric series]\label{fac:geometric_series}
If the following conditions hold
\begin{itemize}
    \item Let $a \in \R$.   
    \item Let $k \in \mathbb{N}$.  
    \item Let $r \in \R$ and $0 < r < 1$.
\end{itemize}

Then, for all $k$, the series which can be expressed in the form of 
    \begin{align*}
        \sum_{i = 0}^k ar^i
    \end{align*}
    is called the geometric series.

    Let $a_0$ denote the value of this series when $k = 0$, namely $a_0 = ar^0 = a$.

    This series is equal to 
    \begin{enumerate}
        \item 
    \begin{align*}
        S_k = \sum_{i = 0}^k ar^i = a_0 \frac{(1 - r^n)}{1 - r},
    \end{align*}
    when $k \not= \infty$, or

    \item
    \begin{align*}
        S_k = \sum_{i = 0}^k ar^i = \frac{a_0}{1 - r},
    \end{align*}
    when $k = \infty$.
    
    \end{enumerate}
\end{fact}

\begin{fact}\label{fac:geo_seires_apply}
If the following conditions hold
\begin{itemize}
    \item Let $\sum_{n = 1}^\infty b_n$ be a series. 
    \item Let $k \in \mathbb{N}$.  
    \item Let $a \in \R$.   
    \item Let $r \in \R$ and $0 < r < 1$.
    \item Let $\sum_{i = 0}^k ar^i$ 
    be a geometric series.
    \item Suppose $\sum_{n = 1}^\infty b_n \leq \sum_{i = 0}^k ar^i$.
\end{itemize}

Then, $\sum_{n = 1}^\infty b_n$ is convergent and is bounded by
    \begin{align*}
        \frac{a_0}{1 - r}.
    \end{align*}
\end{fact}
\begin{proof}

    By Fact~\ref{fac:geometric_series}, we get that the geometric series is convergent, for all $k \in \mathbb{N}$.
    
    Then, $\sum_{n = 1}^\infty b_n$ is convergent by the comparison test.

    We have 
    \begin{align}\label{eq:bound_geo}
        a_0 \frac{(1 - r^n)}{1 - r} \leq \frac{a_0}{1 - r}
    \end{align}
    because for all $0 < r < 1$, we have $(1 - r^n) < 1$.

    Therefore, we get
    \begin{align*}
        \sum_{n = 1}^\infty b_n
        \leq & ~ \sum_{i = 0}^k ar^i \\
        \leq & ~ \frac{a_0}{1 - r}, 
    \end{align*}
    where the first step follows from the assumption in the Fact statement and the second step follows from Eq.~\eqref{eq:bound_geo}.

\end{proof}

\begin{fact}\label{fac:general_p_three_vectors}
If the following conditions hold
\begin{itemize}
    \item $u,v,w\in \mathbb{R}^n$ are three arbitrary unit vectors.
    \item For all $x$ satisfying $0\leq x\leq 1$.
    \item Suppose $1 - x \leq \langle u,w\rangle$.
    \item Suppose $\langle v,  w\rangle =0$.
\end{itemize}

Then
$
\langle u, v \rangle \leq \sqrt{2x - x^2}.
$ 
\end{fact}
\begin{proof}

First, we want to show that
\begin{align}\label{eq:sin_theta(u,w)_leq_(1-x)^2}
    |\sin \theta(u,w)|
    = & ~ \sqrt{1-\cos^2 \theta(u,w)} \notag\\
    = & ~ \sqrt{1-\langle u,w\rangle^2} \notag\\
    \leq & ~ \sqrt{1- (1-x)^2},
\end{align}
where the first step follows from the definition of $\sin \theta(u,w)$ (see Definition~\ref{def:sin_cos_tan}), the second step follows from the definition of $\cos \theta(u,w)$ (see Definition~\ref{def:sin_cos_tan}), and the last step follows from the assumption of this fact. 

Then, we have
\begin{align*}
\langle u, v \rangle
= & ~ \cos \theta(u,v) \\
= & ~ |\cos\theta(u,w) \cos\theta(v,w) - \sin \theta(u,w) \sin\theta(v,w)| \\
\leq & ~ |\cos\theta(u,w) \cos\theta(v,w) + \sin \theta(u,w) \sin\theta(v,w)| \\
\leq & ~ |\cos\theta(u,w) \cos\theta(v,w)| + |\sin \theta(u,w) \sin\theta(v,w)| \\
= & ~ 0 + |\sin \theta(u,w) \sin\theta(v,w)| \\
\leq & ~ |\sin \theta(u,w)| \cdot |\sin\theta(v,w)| \\
\leq & ~ |\sin\theta(u,w)| \\
\leq & ~ \sqrt{1 - (1-x)^2 } \\
= & ~ \sqrt{2x - x^2},
\end{align*}
where the first step follows from the definition of $\cos \theta (u,v)$ (see Definition~\ref{def:sin_cos_tan}), the second step follows from $\cos{(a+b)} = \cos{(a)} \cos{(b)} - \sin{(a)} \sin{(b)}$, the third step follows from simple algebra, the fourth step follows from the triangle inequality, the fifth step follows from $\cos \theta(v,w) = 0$, the sixth step follows from the Cauchy–Schwarz inequality, the seventh step follows from $|\sin \theta(w,v)|\leq 1$, the eighth step follows from Eq.~\eqref{eq:sin_theta(u,w)_leq_(1-x)^2}, and the last step follows from simple algebra.
\end{proof}

\begin{fact}

If the following conditions hold
\begin{itemize}
    \item Let $E \in \mathbb{R}^{n^p}$.
    \item Let $u,v\in \mathbb{R}^n$ be two vectors.
\end{itemize}

Then 
\begin{itemize}
\item $|E(v,u,\cdots,u)| = |v^\top E(I,u,\cdots,u)|$. 
\item $|v^\top E(I,I,u,\cdots, u) w | = |E(v,w,u,\cdots,u)|$
\end{itemize}
\end{fact}
\begin{proof}
It follows 
\begin{align*}
|v^\top E(I,u,\cdots,u)|
= & ~ \left|\sum_{i_1=1}^n v_{i_1}  \cdot \left(\sum_{i_2=1}^n \cdots\sum_{i_p=1}^n E_{i_1,i_2,\cdots,i_p} u_{i_2}\cdots u_{i_p}\right) \right|\\
= & ~ \left|\sum_{i_1=1}^n  \sum_{i_2=1}^n \cdots\sum_{i_p=1}^n E_{i_1,i_2,\cdots,i_p} v_{i_1} u_{i_2}\cdots u_{i_p}  \right| \\
= & ~  |E(v,u,\cdots,u)|,
\end{align*}
where the first step follows from the definition of $E(I,u,\cdots,u)$, the second step follows from the property of summation, and the last step follows from the definition of $E(v,u,\cdots,u)$.
\end{proof}

\begin{fact}\label{fac:bound_u-v}
If the following conditions hold
\begin{itemize}
    \item $u,v$ are two arbitrary unit vectors.
    \item Suppose $\theta(u,v)$ is in the interval $(0,\pi/2)$.
\end{itemize}

Then $\| u-v\|_2 \leq \tan\theta(u,v)$.
\end{fact}
\begin{proof}

Suppose $\theta(u,v)$ is in the interval $ (0,\pi/2)$, so we have 
\begin{align*}
    \cos\theta(u,v)
\end{align*}
is in the interval $(0,1)$.

Let $x = \langle u, v \rangle$. 

Therefore, by the definition of $\cos\theta(u,v)$ (see Definition~\ref{def:sin_cos_tan}), we have 
\begin{align}\label{eq:cos_theta_is_x}
    \cos\theta(u,v) 
    = & ~ \langle u, v \rangle \notag\\
    = & ~ x.
\end{align}
Accordingly, we have
\begin{align}\label{eq:sin_theta_is_1-x^2}
    \sin\theta(u,v) 
    = & ~ \sqrt{1 - \cos^2\theta(u,v)} \notag\\
    = & ~ \sqrt{1 - x^2},
\end{align}
where the first step follows from the definition of $\sin\theta(u,v)$ (see Definition~\ref{def:sin_cos_tan}) and the second step follows from Eq.~\eqref{eq:cos_theta_is_x}.
Moreover, 
\begin{align}\label{eq:u-v_2^2}
    \| u-v\|_2^2 
    = & ~ \| u\|_2^2 + \|v\|_2^2 - 2 \langle u, v \rangle \notag\\
    = & ~ 1 + 1 - 2 x \notag\\
    = & ~ 2 - 2 x,
\end{align}
where the first step follows from simple algebra, the second step follows from the fact that $u$ and $v$ are unit vectors, and the last step follows from simple algebra.
We want to show 
\begin{align*}
    \| u-v\|_2^2 \leq \tan^2 \theta(u,v).
\end{align*}

It suffices to show
\begin{align}\label{eq:2-2x_leq_1-x^2/x^2}
    2 - 2 x 
    \leq & ~ \tan^2 \theta(u,v) \notag\\
    = & ~ \sin^2 \theta(u,v) / \cos^2 \theta(u,v) \notag\\
    \leq & ~ (1- x^2) / x^2,
\end{align}
where the first step follows from Eq.~\eqref{eq:u-v_2^2}, the second step follows from the definition of $\tan \theta(u,v)$ (see Definition~\ref{def:sin_cos_tan}), and the last step follows from combining Eq.~\eqref{eq:cos_theta_is_x} and Eq.~\eqref{eq:sin_theta_is_1-x^2}.

Therefore, it suffices to show 
\begin{align*}
    (1- x^2) / x^2 - (2 - 2 x) \geq 0
\end{align*}
when $x \in (0,1)$.

Let $f : (0, \infty) \to \R$ be defined as
\begin{align*}
    f(x) = (1- x^2) / x^2 - (2 - 2 x).
\end{align*}

Then, the derivative of $f(x)$ is denoted as $f'(x)$, which is as follows
\begin{align*}
    f'(x) = \frac{2x^3 - 2}{x^3}.
\end{align*}

Therefore, when $x = 1$, we have $f'(x) = 0$.

The second derivative of $f$ is 
\begin{align*}
    f''(x) = \frac{6}{x^4}.
\end{align*}

Therefore, 
\begin{align*}
    f''(1) = 6 > 0.
\end{align*}

Thus, $f(1)$ is a local minimum. In other words, when $x \in (0,1)$,
\begin{align*}
    f(x) = (1- x^2) / x^2 - (2 - 2 x) \geq f(1) = 0,
\end{align*}
so Eq.~\eqref{eq:2-2x_leq_1-x^2/x^2} is shown to be true.

Thus, we complete the proof.
\end{proof}

\subsection{More Tensor Facts}\label{sec:appen_prelim:tensor}

In this section, we present more tensor properties.

\begin{fact}[Formal version of Fact~\ref{fac:vj_dot_TIu:informal}]\label{fac:vj_dot_TIu:formal}

If the following conditions hold
\begin{itemize}
    \item Let $p$ be greater than or equal to $3$.
    \item Let $A ^*=\sum_{j=1}^k \lambda_j v_j^{\otimes p} \in \mathbb{R}^{n^p}$ be an orthogonal tensor.
    \item Let $u\in \mathbb{R}^n$ be a vector.
    \item Let $j\in [k]$.
\end{itemize}

Then, we can get 
\begin{align*}
    |v_j^\top A ^*(I,u,\cdots,u)| = \lambda_j |v_j^\top u|^{p-1}.
\end{align*}

\end{fact}
\begin{proof}

For any $j\in [k]$, we have
\begin{align*}
|v_j^\top A ^*(I,u,\cdots, u)| 
 = & ~ \left| \sum_{i=1}^n v_{j,i} A^* (I,u,\cdots,u)_i \right| \\
 = & ~ \left| \sum_{i=1}^n v_{j,i} \sum_{i_2=1}^n \cdots \sum_{i_p=1}^n A ^*_{i,i_2,\cdots,i_p} u_{i_2} \cdots u_{i_p}\right| \\
 = & ~ \left| \sum_{i=1}^n v_{j,i} \sum_{i_2=1}^n \cdots \sum_{i_p=1}^n (\sum_{\ell=1}^n \lambda_\ell v_{\ell,i} v_{\ell,i_2} \cdots v_{\ell,i_p} ) u_{i_2} \cdots u_{i_p}\right| \\
  = & ~ \left| \sum_{\ell=1}^k \lambda_\ell \sum_{i=1}^n v_{j,i}   v_{\ell,i} \sum_{i_2=1}^n \cdots \sum_{i_p=1}^n ( v_{\ell,i_2} \cdots v_{\ell,i_p} ) u_{i_2} \cdots u_{i_p}\right| \\
   = & ~ \left| \lambda_j     \sum_{i_2=1}^n \cdots \sum_{i_p=1}^n ( v_{j,i_2} \cdots v_{j,i_p} ) u_{i_2} \cdots u_{i_p}\right| \\
= & ~ \lambda_j |v_j^\top u|^{p-1},
\end{align*}
where the first step follows from the definition of vector norm, the second step follows from the decomposition of $A^*$ by its definition, the third step follows from the definition of $A^*$, the fourth step follows from reordering the summations, the fifth step follows from taking summations over $\ell$, and the sixth step follows from simple algebra.
\end{proof}

\begin{fact}[Formal version of Fact~\ref{fac:diff_two_tensor_l2:informal}]\label{fac:diff_two_tensor_l2:formal}
If the following conditions hold
\begin{itemize}
    \item Let $p\geq 3$.
    \item $x,y,u,v\in \mathbb{R}^n$ are four arbitrary unit vectors.
    \item Let $j\in \{0,1,\cdots,p-2\}$.
\end{itemize}

Then, we can get
\begin{align}\label{eq:equal_x-y:formal}
 \| [x\otimes v^{\otimes (p-1)}](I,u,\cdots,u) 
- [y\otimes v^{\otimes (p-1)}](I,u,\cdots,u) \|_2
= |\langle u, v\rangle|^{p-1} \cdot \| x - y\|_2
\end{align}
and
\begin{align}\label{eq:leq_x-y:formal}
 & ~ \| [  v^{\otimes (1+j)} \otimes x \otimes v^{\otimes (p-2-j)}](I,u,\cdots,u) - [v^{\otimes (1+j)} \otimes y \otimes v^{\otimes (p-2-j)}](I,u,\cdots,u) \|_2 \notag\\
\leq & ~ |\langle u,v \rangle|^{p-2} \cdot \| x - y \|_2.
\end{align} 
\end{fact}
\begin{proof}
To show Eq.~\eqref{eq:equal_x-y:formal}, let's analyze the $i$-th entry of the vector 
\begin{align*}
    [x\otimes v^{\otimes (p-1)}](I,u,\cdots,u)\in \mathbb{R}^n,
\end{align*}
which can be written as
\begin{align}\label{eq:x_otimes_v}
x_i \sum_{i_2=1}^n \cdots \sum_{i_p=1}^n v_{i_2} \cdots v_{i_p} u_{i_2} \cdots u_{i_p}
= & ~ x_i \sum_{i_2=1}^n v_{i_2} u_{i_2} \cdots \sum_{i_p=1}^n  v_{i_p} u_{i_p} \notag\\
= & ~ x_i \langle v, u \rangle^{p-1},
\end{align}
where the first step follows from the property of summation and the second step follows from the definition of the inner product. 

In this part, for simplicity, we define
\begin{align*}
    \mathrm{LHS} := \| [x\otimes v^{\otimes (p-1)}](I,u,\cdots,u) - [y\otimes v^{\otimes (p-1)}](I,u,\cdots,u) \|_2.
\end{align*}
By Eq.~\eqref{eq:x_otimes_v}, we have
\begin{align*}
    \mathrm{LHS} = \|x_i \langle v, u \rangle^{p-1} - y_i \langle v, u \rangle^{p-1}\|_2.
\end{align*}
Thus, we get
\begin{align*}
\mathrm{LHS}^2
= & ~ \sum_{i=1}^n (x_i \langle v, u\rangle^{p-1}  - y_i \langle v, u\rangle^{p-1})^2\\
= & ~ \sum_{i=1}^n ((x_i - y_i) (\langle v, u\rangle^{p-1}))^2\\
= & ~ \sum_{i=1}^n (x_i - y_i)^2 (\langle v, u\rangle^{p-1})^2\\
= & ~ \langle v, u\rangle^{2(p-1)} \sum_{i=1}^n (x_i - y_i)^2 \\
= & ~ \| x - y\|_2^2 \cdot |\langle v,u \rangle|^{2(p-1)},
\end{align*}
where the first step follows from the definition of $\|\cdot\|_2$, the second step follows from simple algebra, the third step follows from simple algebra, the fourth step follows from the fact that $i$ is not contained in $\langle v, u\rangle^{2(p-1)}$, and the last step follows from the definition of $\| \cdot \|_2$.

To show Eq.~\eqref{eq:leq_x-y:formal}, first, we want to show
\begin{align}\label{eq:x-y_u}
    |\langle x-y,u\rangle| 
    \leq & ~ \| x- y \|_2 \| u \|_2 \notag \\
    \leq & ~  \| x - y \|_2,
\end{align}
where the first step follows from the Cauchy–Schwarz inequality and the second step follows from the fact that $u$
is a unit vector so that $\| u \|_2=1$.

Then, we analyze the $i$-th entry of the vector 
\begin{align*}
    [v^{\otimes (1+j)}\otimes x\otimes v^{\otimes (p-2-j)}](I,u,\cdots,u)\in \mathbb{R}^n,
\end{align*}
which is equivalent to  
\begin{align}\label{eq:v^otimes_1+j}
    v_i \langle x, u \rangle \cdot \langle v, u \rangle^{p-2}.
\end{align}

In this part, we define
\begin{align*}
    \mathrm{LHS} := \| [  v^{\otimes (1+j)} \otimes x \otimes v^{\otimes (p-2-j)}](I,u,\cdots,u) - [v^{\otimes (1+j)} \otimes y \otimes v^{\otimes (p-2-j)}](I,u,\cdots,u) \|_2.
\end{align*}
Therefore, based on Eq.~\eqref{eq:v^otimes_1+j}, we get
\begin{align*}
    \mathrm{LHS} = \| v_i \langle x, u \rangle \cdot \langle v, u \rangle^{p-2} - v_i \langle y, u \rangle \cdot \langle v, u \rangle^{p-2} \|_2
\end{align*}
Thus, we have
\begin{align*}
\mathrm{LHS}^2
= & ~ \sum_{i=1}^n (v_i \langle x, u \rangle \cdot \langle v,u \rangle^{p-2} - v_i \langle y, u \rangle \cdot \langle v,u \rangle^{p-2} )^2\\
= & ~ \sum_{i=1}^n ((v_i \langle x, u \rangle - v_i \langle y, u \rangle) \cdot \langle v,u \rangle^{p-2} )^2\\
= & ~ \sum_{i=1}^n ((v_i \langle x, u \rangle - v_i \langle y, u \rangle)^2 \cdot \langle v,u \rangle^{2(p-2)})\\
= & ~ \langle v,u \rangle^{2(p-2)} \cdot \sum_{i=1}^n (v_i \langle x, u \rangle - v_i \langle y, u \rangle)^2 \\
= & ~ \langle v,u \rangle^{2(p-2)} \cdot \sum_{i=1}^n (v_i (\langle x, u \rangle - \langle y, u \rangle))^2 \\
= & ~ \langle v,u \rangle^{2(p-2)} \cdot \sum_{i=1}^n (v_i \langle x - y, u \rangle)^2 \\
= & ~ \langle v,u \rangle^{2(p-2)} \cdot \sum_{i=1}^n ((v_i^2)( \langle x - y, u \rangle^2)) \\
= & ~ \langle x-y, u\rangle^2 \cdot \langle v, u \rangle^{2(p-2)} \sum_{i=1}^n (v_i)^2\\
= & ~ \langle x-y, u\rangle^2 \cdot \langle v, u \rangle^{2(p-2)}\\
\leq & ~ \| x - y\|_2^2 \cdot \langle v,u\rangle^{2(p-2)},
\end{align*}
where the first step follows from the definition of $\| \cdot \|_2$, the second step follows from simple algebra, the third step follows from $(ab)^2 = a^2 b^2$, the fourth step follows from the fact that $i$ is not contained in $\langle v,u \rangle^{2(p-2)}$, the fifth step follows from simple algebra, the sixth step follows from the linearity property of the inner product, the seventh step follows from $(ab)^2 = a^2 b^2$, the eighth step follows from the fact that $i$ is not contained in $\langle x-y, u\rangle^2$, the ninth step follows from the fact that $v$ is a unit vector, and the last step follows from Eq.~\eqref{eq:x-y_u}.
\end{proof}

\begin{fact}[Formal version of Fact~\ref{fac:bounding_E_p:informal}]\label{fac:bounding_E_p:formal}
If the following conditions hold
\begin{itemize}
    \item Let $E \in \mathbb{R}^{n^p}$ be an arbitrary tensor.
    \item $u,v\in \mathbb{R}^n$ are two unit vectors.
\end{itemize}

Then,
\begin{enumerate}
\item $| E(u,v,\cdots,v) | \leq \| E \|$.
\item $\| E(I,v,\dots,v) \|_2 \leq \sqrt{n} \| E\|$.
\end{enumerate}
\end{fact}
\begin{proof}
Part 1 trivially follows from the definition of $\| E\|$.  

For part 2, we define a unit vector $w \in \R^n$ 
to be $(1/\sqrt{n}, \cdots, 1/ \sqrt{n})$,
\begin{align*}
\| E(I,v,\dots,v) \|_2^2 
= & ~ \sum_{i_1=1}^n \left( \sum_{i_2=1}^n \cdots \sum_{i_p=1}^n E_{i_1,i_2,\cdots,i_p} v_{i_2} \cdots v_{i_p} \right)^2 \\
= & ~ n \sum_{i_1=1}^n \left( \sum_{i_2=1}^n \cdots \sum_{i_p=1}^n E_{i_1,i_2,\cdots,i_p} w_{i_1} v_{i_2} \cdots v_{i_p} \right)^2 \\
\leq & ~ n \| E\|^2,
\end{align*}
where the first step follows from the definition of $E(I,v,\dots,v)$, the second step follows from our definition for $w$, and the last step follows from $n \geq 1$ .

This result implies $\| E(I,v,\cdots,v) \|_2 \leq \sqrt{n} \| E\|$.
\end{proof}

\begin{fact}[Formal version of Fact~\ref{fac:V_dot_T:informal}]\label{fac:V_dot_T:formal}
If the following conditions hold
\begin{itemize}
    \item $v_1, v_2, \cdots, v_n$ is an orthonormal basis.
    \item $V = (v_2,\cdots,v_n) \in \mathbb{R}^{n\times (n-1)}$.
    \item Let $A ^* = \sum_{i=1}^k \lambda_i v_i^{\otimes p}$.
    \item $u\in \mathbb{R}^n$ is an arbitrary vector.
\end{itemize}

Then, we have 
\begin{align*}
    \| V^\top A ^*(I,u,\cdots, u) \|_2^2 = \sum_{j=2}^k \lambda_j^2 |v_j^\top u|^{2(p-1)}.
\end{align*}
\end{fact}
\begin{proof}
We have
\begin{align*}
 \| V^\top A ^*(I,u,\cdots, u) \|_2^2 
 = & ~ \sum_{j=2}^k  |v_j^\top A ^*(I,u,\cdots, u)|^2 \\
 = & ~ \sum_{j=2}^k  (\lambda_j |v_j^\top u|^{p-1})^2 \\
 = & ~ \sum_{j=2}^k \lambda_j^2 |v_j^\top u|^{2(p-1)},
\end{align*}
where the first step follows from the definition of $\ell_2$ norm, the second step follows from Fact~\ref{fac:vj_dot_TIu:formal}, and the last step follows from $(ab)^2 = a^2 b^2$.
\end{proof}

%% file: app_analysis.tex
\section{More Analysis}

In Section~\ref{sub:more_analysis:part_1}, we give the proof to the first part of Theorem~\ref{thm:general_p_bound_EIu}. In Section~\ref{sub:more_analysis:part_2}, we give the proof to the second part of Theorem~\ref{thm:general_p_bound_EIu}. In Section~\ref{sub:more_analysis:part_3}, we give the proof to the third part of Theorem~\ref{thm:general_p_bound_EIu}. In Section~\ref{sub:more_analysis:epsilon}, we prove that a few terms are upper-bounded.

\label{sec:more_analysis}

\subsection{Part 1 of Theorem~\ref{thm:general_p_bound_EIu}}
\label{sub:more_analysis:part_1}

In this section, we present the proof of the first part of Theorem~\ref{thm:general_p_bound_EIu}.

For convenient, we first create some definitions for this section
\begin{definition}\label{def:B_1_B_2}
We define $B_1 \in \R$ and $B_2 \in \R$ as follows 
\begin{align*}
B_1 : = \frac{1}{1-|v_1^\top \wt{E}_{u_t}| /(\lambda_1 |v_1^\top u_t|^{p-1})}
\end{align*}
We define
\begin{align*}
B_2 : = \frac{\| V^\top \wt{E}_{u_t}\|_2 }{\lambda_1 |v_1^\top u_t|^{p-1}}
\end{align*}
\end{definition}

\begin{lemma}[Part 1 of Theorem~\ref{thm:general_p_bound_EIu}]
If the following conditions hold
\begin{itemize}
    \item Let everything be defined as in Theorem~\ref{thm:general_p_bound_EIu}. 
    \item Suppose that all of the assumptions in Theorem~\ref{thm:general_p_bound_EIu} hold.
\end{itemize}

Then, Eq.~\eqref{eq:part_1_of_thm} hold.
\end{lemma}

\begin{proof}

{\bf Proof of Part 1.}

$V = (v_2, \cdots,v_k,\cdots, v_n) \in \mathbb{R}^{n\times (n-1)}$ is an orthonormal basis and is the complement of $v_1$.

Also, $\wt{E}_{u_t} = \wt{E}(I,u_t,\cdots, u_t)\in \mathbb{R}^n$. 

$\tan \theta(v_1,u_{t+1})$'s upper bound is provided as follows:
\begin{align}\label{eq:rewrite_tan_theta_v_1_u_t+1}
 \tan\theta(v_1, u_{t+1})
 = & ~ \tan \theta \left(v_1, \frac{A (I,u_t, \cdots, u_t)}{\| A (I,u_t, \cdots, u_t) \|_2 }\right) \notag \\
= & ~ \tan \theta (v_1, A (I,u_t, \cdots, u_t)  ) \notag \\
= & ~ \tan \theta (v_1, A ^*(I,u_t, \cdots, u_t) +\wt{E}(I,u_t, \cdots, u_t) ) \notag \\
= & ~ \tan \theta (v_1, A ^*(I,u_t, \cdots, u_t) + \wt{E}_{u_t} ) \notag \\
= & ~ \frac{ \sin \theta (v_1, A ^*(I,u_t, \cdots, u_t) + \wt{E}_{u_t} ) }{ \cos \theta (v_1, A ^*(I,u_t, \cdots, u_t) + \wt{E}_{u_t} ) } \notag \\
= & ~ \frac{\| V^\top [ A ^*(I,u_t,\cdots,u_t) + \wt{E}_{u_t} ] \|_2}{ |v_1^\top [A ^*(I,u_t,\cdots,u_t) + \wt{E}_{u_t} ] | } \notag \\
\leq & ~ \frac{ \| V^\top A ^*(I,u_t,\cdots, u_t) \|_2 + \| V^\top \wt{E}_{u_t} \|_2 }{ |v_1^\top A ^*(I, u_t, \cdots, u_t ) | - |v_1^\top \wt{E}_{u_t}| },
\end{align}
where the first step follows from the definition of $u_{t+1}$, the second step follows from the definition of angle, the third step follows from $A = A^* + \wt{E} \in \mathbb{R}^{n^p}$, the fourth step follows from $\wt{E}_{u_t} := \wt{E}(I,u_t,\cdots,u_t)\in \mathbb{R}^n$, the fifth step follows from the definition of $\tan \theta$, the sixth step follows from $\sin$ and $\cos$, and the seventh step follows from the triangle inequality.

Using Fact~\ref{fac:V_dot_T:formal}, we can get 
\begin{align}\label{eq:bound_VTIut}
\| V^\top A ^*(I,u_t,\cdots, u_t) \|_2^2
= & ~ \sum_{j=2}^k \lambda_j^2 |v_j^\top u_t|^{2(p-1)} \notag\\
\leq & ~ \left(\max_{j \in [k]\backslash\{1\} } |\lambda_j|^2 |v_j^\top u_t|^{2(p-2)} \right) \cdot \left(\sum_{j=2}^k |v_j^\top u_t|^2\right),
\end{align}
where the second step follows from $\sum_{i} a_i b_i \leq ( \max_i a_i) \cdot \sum_i b_i $ for all $a,b \in R^n_{\geq 0}$.

Putting it all together, we have
\begin{align*}
\tan \theta(v_1, u_{t+1}) 
\leq & ~ \frac{\| V^\top A ^*(I,u_t,\cdots,u_t) \|_2 + \|V^\top \wt{E}_{u_t} \|_2}{ |v_1^\top A ^*(I,u_t,\cdots,u_t)| - |v_1^\top \wt{E}_{u_t} | } \\
\leq & ~ \tan \theta(v_1,u_t) \cdot \frac{ ( \| V^\top A ^*(I,u_t,\cdots,u_t) \|_2  + \|V^\top \wt{E}_{u_t} \|_2 )/ \| V^\top u_t\|_2 }{ |v_1^\top A ^*(I,u_t,\cdots,u_t)|/|v_1^\top u_t| - |v_1^\top \wt{E}_{u_t} |/ |v_1^\top u_t| } \\ 
\leq & ~ \tan \theta(v_1,u_t) \cdot \frac{ \underset{j\in[k]\backslash\{1\} }{\max} \lambda_j |v_j^\top u_t|^{p-2} + \|V^\top \wt{E}_{u_t} \|_2/ \| V^\top u_t\|_2 }{ |v_1^\top A ^*(I,u_t,\cdots,u_t)|/|v_1^\top u_t| - |v_1^\top \wt{E}_{u_t} |/ |v_1^\top u_t| } \\
\leq & ~ \tan \theta(v_1,u_t) \cdot \frac{ \underset{j\in[k]\backslash\{1\} }{\max} \lambda_j |v_j^\top u_t|^{p-2} + \|V^\top \wt{E}_{u_t} \|_2/ \| V^\top u_t\|_2 }{ \lambda_1 |v_1^\top u_t|^{p-2} - |v_1^\top \wt{E}_{u_t} |/ |v_1^\top u_t| } \\ 
\leq & ~ \tan \theta(v_1,u_t) \cdot \frac{ (1/4) \lambda_1 |v_1^\top u_t|^{p-2} + \|V^\top \wt{E}_{u_t} \|_2/ \| V^\top u_t\|_2 }{ \lambda_1 |v_1^\top u_t|^{p-2} - |v_1^\top \wt{E}_{u_t} |/ |v_1^\top u_t| } \\ 
\leq & ~ \tan \theta(v_1,u_t) \cdot (1/4) \cdot \underbrace{\frac{1}{1-|v_1^\top \wt{E}_{u_t}| /(\lambda_1 |v_1^\top u_t|^{p-1}) } }_{B_1}  \\
+ &~   \underbrace{\frac{1}{1-|v_1^\top \wt{E}_{u_t}| /(\lambda_1 |v_1^\top u_t|^{p-1}) } }_{B_1} \cdot \underbrace{ \frac{\| V^\top \wt{E}_{u_t}\|_2 }{\lambda_1 |v_1^\top u_t|^{p-1}} }_{B_2} \\
= & ~ \tan\theta(v_1,u_t) \cdot (1/4) \cdot B_1 + B_1\cdot B_2,
\end{align*}
where the 1st step comes from Eq.~\eqref{eq:rewrite_tan_theta_v_1_u_t+1}, the 2nd step is by $\tan\theta(v_1,u_t) = \frac{\| V^\top u_t \|_2}{ |v_1^\top u_t| }$, the 3rd step is because of Equation~(\ref{eq:bound_VTIut}), the 4th step follows from Fact~\ref{fac:vj_dot_TIu:formal}, the 5th step follows from Part~2~of~Theorem~\ref{thm:general_p_bound_EIu}, the 6th step follows from simple algebra, and the 7th step follows from the definition of $B_1$ and $B_2$.

We show  
\begin{claim}
For any $t\in [T]$, we have
\begin{align*}
    |v_1^\top u_0| \leq |v_1^\top u_t|.
\end{align*}

\end{claim}
\begin{proof}
Based on the assumption from the induction hypothesis, we consider the existence of a sufficiently small constant $c$ being greater than $0$ satisfying
\begin{align}\label{eq:recursive}
    \tan\theta(v_1,u_t) \leq 0.8 \tan \theta(v_1,u_{t-1}) + c.
\end{align}

Therefore, we can get
\begin{align*}
\tan\theta(v_1,u_t)
\leq & ~ 0.8 \cdot (0.8\tan\theta(v_1,u_{t-1}) + c ) + c \\
\leq & ~ 0.8^t \cdot \tan\theta(v_1,u_{0}) + c \sum_{j=0}^{t-1} 0.8^j \\
\leq & ~ 0.8^t \cdot \tan\theta(v_1,u_{0}) +  5 c \\
\leq & ~ \tan \theta(v_1,u_0),
\end{align*}
where the first step follows from applying Eq.~\eqref{eq:recursive} recursively twice, the second step follows from applying Eq.~\eqref{eq:recursive} recursively for $t+1$ times, the third step follows from $\sum_{j=0}^{\infty} 0.8^j \leq 5$, and the last step follows from $\tan\theta(v_1,u_0) = \Omega(1)$.

This result shows 
\begin{align*}
    \theta(v_1,u_t) \leq \theta(v_1,u_0),
\end{align*}
so
\begin{align*}
    |v_1^\top u_t| = \cos\theta(v_1,u_t) \geq \cos \theta(v_1,u_0) = |v_1^\top u_0|.
\end{align*}

\end{proof}
Therefore, $B_1$ and $B_2$ has upper bounds.
\begin{claim}
$B_1$ is smaller than or equal to $1.1$.
\end{claim}
\begin{proof}
Let's consider 
\begin{align*}
    | \wt{E}(v_j,u_t,\cdots,u_t)| 
    = & ~ |v_j^\top \wt{E}(I,u_t,\cdots,u_t)| \\
    = & ~ |v_j^\top \wt{E}_{u_t}|.
\end{align*}

Since 
\begin{align*}
| \wt{E}(v_j,u_t,\cdots,u_t)| \leq & ~4  \epsilon /\sqrt{n} \\
 \leq & ~ 4c\lambda_1 / n^{(p-1)/2}\\
 \leq & ~ 4c \lambda_1 |v_1^\top u_0|^{p-1},
\end{align*}
where the first step follows from the constraint on $\wt{E}$ in Theorem~\ref{thm:general_p_bound_EIu},
the second step follows from $\epsilon \leq c\lambda_1 /n^{(p-2)/2}$, and the third step follows from $|v_1^\top u_0| \geq 1/\sqrt{n}$.

Correspondingly, if $c$ can be chosen to be small enough, i.e., $c$ is smaller than $\frac{1}{40}$, using
\begin{align*}
    |v_1^\top \wt{E}_{u_t}| \leq \lambda_1 |v_1^\top u_0|^{p-1} /10
\end{align*}
and 
\begin{align*}
    | v_1^\top u_0 | \leq |v_1^\top u_t|,
\end{align*}
then
\begin{align}\label{eq:v_1_top_E_ut}
|v_1^\top \wt{E}_{u_t}|
\leq & ~ \lambda_1 |v_1^\top u_0|^{p-1} /10 \notag\\
\leq & ~ \lambda_1 |v_1^\top u_t|^{p-1} /10.
\end{align}

As a result, 
\begin{align*}
    B_1 \leq \frac{1}{1-1/11} = 1.1.
\end{align*}

\end{proof}

Next, we bound $B_2$. Let's consider two different cases. The first one is
\begin{align*}
    |v_1^\top u_t| \leq 1-\frac{1}{c_0^2 p^2 k^2}
\end{align*}
and the other is 
\begin{align*}
    |v_1^\top u_t|> 1-\frac{1}{c_0^2 p^2 k^2}.
\end{align*}

If 
\begin{align*}
    |v_1^\top u_t| \leq (1-\frac{1}{c_0^2 p^2 k^2}),
\end{align*}
we have
\begin{align*}
B_2 = & ~ \frac{\| V^\top\wt{E}_{u_t}\|_2 }{ \lambda_1 | v_1^\top u_t|^{p-1} }\\
= & ~ \frac{\sqrt{1-|v_1^\top u_t|^2 }}{ |v_1^\top u_t|} \cdot \frac{\| V^\top \wt{E}_{u_t}\|_2 }{ \lambda_1 | v_1^\top u_t|^{p-2} \sqrt{1-|v_1^\top u_t|^2} } \\
= & ~ \tan\theta( v_1, u_t) \cdot \frac{\| V^\top \wt{E}_{u_t}\|_2 }{ \lambda_1 | v_1^\top u_t|^{p-2} \sqrt{1-|v_1^\top u_t|^2} } \\
\leq & ~ \tan\theta( v_1, u_t) \cdot \frac{\| \wt{E}_{u_t}\|_2 }{ \lambda_1 | v_1^\top u_t|^{p-2} \sqrt{1-|v_1^\top u_t|^2} }\\
\leq & ~ \tan\theta( v_1, u_t) \cdot \frac{ c_0pk \| \wt{E}_{u_t}\|_2 }{ \lambda_1 | v_1^\top u_t|^{p-2}  },
\end{align*}
where the first step comes from the definition of $B_2$ (see Definition~\ref{def:B_1_B_2}),
the second step follows from splitting the term,
the third step follows from $\tan\theta( v_1, u_t)  = \frac{\sqrt{1-|v_1^\top u_t|^2 }}{ |v_1^\top u_t|}$,
the fourth step follows that $\| V^\top\wt{E}_{u_t} \|_2 \leq \|\wt{E}_{u_t} \|_2$,
and the last step follows from $1/\sqrt{1-|v_1^\top u_t|^2} \leq c_0pk$.

We need to bound 
\begin{align*}
    \frac{ c_0pk \| \wt{E}_{u_t}\|_2 }{ \lambda_1 | v_1^\top u_t|^{p-2}  }.
\end{align*}

Here, we can get 
\begin{align*}
   \lambda_1 |v_1^\top u_t|^{p-2}  \geq \lambda_1 /(n^{(p-2)/2}  ).
 \end{align*} 
 On the other hand, utilizing Part 1 of Lemma~\ref{lem:kappa_phi} and the given assumptions about $\ov{E}$ and $E$, we obtain $\| \wt{E}_{u_t} \|_2$
 \begin{align*}
c_0 p k  \| \wt{E}_{u_t} \|_2 \leq c_0 p k \cdot 4 p \epsilon.
 \end{align*}
 Consequently, whenever we have small enough $\epsilon$ satisfying
 \begin{align*}
     \epsilon \leq \lambda_1 /( n^{(p-2)/2} \cdot 40 \cdot c_0 p^2 k),
 \end{align*}
 then 
 \begin{align*}
     B_2 \leq 0.1 \tan\theta(v_1,u_t).
 \end{align*}

If 
\begin{align*}
    |v_1^\top u_t|> 1-\frac{1}{c_0^2 k^2 p^2},
\end{align*}
then we have
\begin{align*}
B_2 = & ~ \frac{\| V^\top \wt{E}_{u_t}\|_2 }{ \lambda_1 | v_1^\top u_t|^{p-1} }\\
\leq & ~ \frac{\| V^\top \wt{E}_{u_t}\|_2 }{ \lambda_1 (1-\frac{1}{c_0^2p^2 k^2})^{p-1} }\\
\leq & ~ 3 \| V^\top \wt{E}_{u_t}\|_2 / \lambda_1\\
\leq & ~ 3 \| \wt{E}_{u_t}\|_2 /\lambda_1 
\end{align*}
where the first step follows from the definition of $B_2$, the second step follows from $|v_1^\top u_t|> 1-\frac{1}{c_0^2p^2 k^2}$, the third step follows from 
\begin{align*}
    1/(1-\frac{1}{c_0^2p^2k^2})^{p-1}\leq 3, \forall p\geq 3,k\geq 1, c_0\geq 1,
\end{align*}
and the last step follows from $\| V^\top\wt{E}_{u_t} \|_2 \leq \|\wt{E}_{u_t} \|_2$.

By Part 1 of Corollary~\ref{cor:kappa_phi}, we have $\| \wh{E}_{u_t} \|_2 \leq 4\epsilon /c_0$. By what we have assumed on $E$ and $\ov{E}$, $\| {E}_{u_t} \|_2 \leq \epsilon/c_0$ and $\| \ov{E}_{u_t} \|_2 \leq \epsilon /c_0$, which completes the proof of $B_2  
\leq 18 \epsilon /(c_0\lambda_1)$.

\end{proof}

\subsection{Part 2 of Theorem~\ref{thm:general_p_bound_EIu}}
\label{sub:more_analysis:part_2}

In this section, we present the proof of the second part of Theorem~\ref{thm:general_p_bound_EIu}.

\begin{lemma}[Part 2 of Theorem~\ref{thm:general_p_bound_EIu}]
If the following conditions hold
\begin{itemize}
    \item Let everything be defined as in Theorem~\ref{thm:general_p_bound_EIu}. 
    \item Suppose that all of the assumptions in Theorem~\ref{thm:general_p_bound_EIu} hold.
\end{itemize}

Then, Eq.~\eqref{eq:part_2_of_thm} hold.
\end{lemma}

\begin{proof}

{\bf Proof of Part 2.}

Let $j$ be an arbitrary element in $[k]\backslash\{1\}$. 

Then, there exists an lower bound for $\frac{ |v_1^\top u_{t+1}|}{ |v_j^\top u_{t+1}|}$,
\begin{align}\label{eq:v1ut1_vjut1}
\frac{ |v_1^\top u_{t+1}|}{ |v_j^\top u_{t+1}|}
= & ~ \frac{  |v_1^\top [A ^*(I,u_t,\cdots,u_t) +\wt{E}_{u_t}] |}{ |v_j^\top [A ^*(I,u_t,\cdots,u_t) + \wt{E}_{u_t}]|} \notag \\
\geq & ~ \frac{   |v_1^\top A ^*(I,u_t,\cdots,u_t) | - |v_1^\top \wt{E}_{u_t} | }{    |v_j^\top A ^*(I,u_t,\cdots,u_t) |+ |v_j^\top \wt{E}_{u_t}| } \notag\\
\geq & ~ \frac{ \lambda_1 |v_1^\top u_t|^{p-1} - |v_1^\top \wt{E}_{u_t} | }{   \lambda_j |v_j^\top u_t|^{p-1} + |v_j^\top \wt{E}_{u_t}|  } \notag\\
\geq & ~\frac{  \lambda_1 |v_1^\top u_t|^{p-1} - \frac{1}{10} \lambda_1 |v_1^\top u_t|^{p-1}  }{   \lambda_j |v_j^\top u_t|^{p-1} + \frac{1}{10}\lambda_1 |v_1^\top u_t|^{p-1} } \notag\\
\geq & ~\frac{  \lambda_1 |v_1^\top u_t|^{p-1} - \frac{1}{10} \lambda_1 |v_1^\top u_t|^{p-1} }{   \frac{1}{4} \lambda_1 |v_1^\top u_t|^{p-2} |v_j^\top u_t| + \frac{1}{10}\lambda_1 |v_1^\top u_t|^{p-1} } \notag \\
= & ~ \frac{ \frac{9}{10} |v_1^\top u_t|}{ \frac{1}{4} |v_j^\top u_t| + \frac{1}{10} |v_1^\top u_t|  } 
\end{align}
where the first step follows from the definition of $u_{t+1}$ (see the statement in Theorem~\ref{thm:general_p_bound_EIu}), the second step follows from the triangle inequality, the third step follows from Fact~\ref{fac:vj_dot_TIu:formal}, the fourth step follows from Eq.~\eqref{eq:v_1_top_E_ut},
the fifth step follows from Part 1 $\lambda_j |v_j^\top u_t|^{p-2} \leq \frac{1}{4} \lambda_1 |v_1^\top u_t|^{p-2}$, and the last step follows from simple algebra.

If 
\begin{align}\label{eq:v_j_top_u_t_<}
    |v_j^\top u_t| < |v_1^\top u_t|,
\end{align}
then 
\begin{align}\label{eq:lambda_2_p-2}
\frac{\lambda_1 |v_1^\top u_{t+1}|^{p-2} }{\lambda_j |v_j^\top u_{t+1}|^{p-2}} 
= & ~ \frac{\lambda_1}{\lambda_j} \left( \frac{|v_1^\top u_{t+1}|}{|v_j^\top u_{t+1}|} \right)^{p-2} \notag\\
\geq & ~ \frac{\lambda_1}{\lambda_j} \left(\frac{ \frac{9}{10} |v_1^\top u_t|}{ \frac{1}{4} |v_j^\top u_t| + \frac{1}{10} |v_1^\top u_t|  } \right)^{p-2} \notag\\
\geq & ~ \frac{\lambda_1}{\lambda_j} \left(\frac{ \frac{9}{10} |v_j^\top u_t|}{ \frac{1}{4} |v_j^\top u_t| + \frac{1}{10} |v_j^\top u_t|  } \right)^{p-2} \notag\\
= & ~ \frac{\lambda_1}{\lambda_j} \left(\frac{ \frac{9}{10}}{ \frac{1}{4} + \frac{1}{10} } \right)^{p-2} \notag\\
\geq & ~ \frac{\lambda_1}{\lambda_j} 2^{p-2},
\end{align}
where the first step follows from $\frac{a^x}{b^x} = \left( \frac{a}{b} \right)^x$, the second step follows from Eq.~\eqref{eq:v1ut1_vjut1}, the third step follows from Eq.~\eqref{eq:v_j_top_u_t_<}, the fourth step follows from simple algebra, the last step follows from simple algebra. 

The final step is a consequence of the fact that $p$ is greater than or equal to $4$. For the case of $p$ being equal to $3$, we utilize a better analysis which is similar to the proof of \cite{wa16}'s Lemma C.2. Therefore, this approach is applicable for any $p\geq 3$.

If 
\begin{align}\label{eq:v_j_top_u_t_>}
    |v_j^\top u_t| \geq |v_1^\top u_t|,
\end{align}
then 
\begin{align*}
\frac{\lambda_1 |v_1^\top u_{t+1}|^{p-2} }{\lambda_j |v_j^\top u_{t+1}|^{p-2}}
\geq & ~ \frac{\lambda_1}{\lambda_j} \left(\frac{ \frac{9}{10} |v_1^\top u_t|}{ \frac{1}{4} |v_j^\top u_t| + \frac{1}{10} |v_1^\top u_t|  } \right)^{p-2}\\
= & ~ \frac{\lambda_1}{\lambda_j} \left(\frac{ \frac{9}{10} |v_1^\top u_t| |v_j^\top u_t|}{ \left(\frac{1}{4} |v_j^\top u_t| + \frac{1}{10} |v_1^\top u_t| \right)|v_1^\top u_t| } \frac{|v_1^\top u_t| }{ |v_j^\top u_t|} \right)^{p-2}\\
= & ~ \frac{\lambda_1}{\lambda_j} \left(\frac{ \frac{9}{10} |v_1^\top u_t| |v_j^\top u_t|}{ \left(\frac{1}{4} |v_j^\top u_t| + \frac{1}{10} |v_1^\top u_t| \right)|v_1^\top u_t| }\right)^{p-2}  \left(\frac{|v_1^\top u_t| }{ |v_j^\top u_t|} \right)^{p-2}\\
= & ~ \frac{\lambda_1}{\lambda_j} \left(\frac{ \frac{9}{10}  |v_j^\top u_t|}{ \frac{1}{4} |v_j^\top u_t| + \frac{1}{10} |v_1^\top u_t|  }\right)^{p-2}  \left(\frac{|v_1^\top u_t| }{ |v_j^\top u_t|} \right)^{p-2}\\
\geq & ~ \frac{\lambda_1}{\lambda_j} \left(\frac{ \frac{9}{10}  |v_j^\top u_t|}{ \frac{1}{4} |v_j^\top u_t| + \frac{1}{10} |v_j^\top u_t|  }\right)^{p-2}  \left(\frac{|v_1^\top u_t| }{ |v_j^\top u_t|} \right)^{p-2}\\
\geq & ~ \frac{\lambda_1}{\lambda_j}  \frac{|v_1^\top u_t|^{p-2} }{ |v_j^\top u_t|^{p-2}} \cdot \left(\frac{ \frac{9}{10}  |v_j^\top u_t|}{ \frac{1}{4} |v_j^\top u_t| + \frac{1}{10} |v_j^\top u_t|  }\right)^{p-2} \\
\geq & ~ \frac{\lambda_1}{\lambda_j}  \frac{|v_1^\top u_t|^{p-2} }{ |v_j^\top u_t|^{p-2}} \cdot 2^{p-2} \\
\geq & ~ 4 \cdot 2^{p-2},
\end{align*}
where the first step follows from the second step of Eq.~\eqref{eq:lambda_2_p-2}, the second step follows from simple algebra, the third step follows from $(ab)^2 = a^2 b^2$, the fourth step follows from simple algebra, the fifth step follows from Eq.~\eqref{eq:v_j_top_u_t_>}, the sixth step follows from $\left( \frac{a}{b} \right)^2 = \frac{a^2}{b^2}$, the seventh step follows from the relationship between the third step and the last step of Eq.~\eqref{eq:lambda_2_p-2}, and the last step follows from Part 1.

\end{proof}

\subsection{Part 3 of Theorem~\ref{thm:general_p_bound_EIu}}

In this section, we present the proof of the third part of Theorem~\ref{thm:general_p_bound_EIu}.

\label{sub:more_analysis:part_3}

\begin{definition}
We define $B_3 \in \R$ and $B_4 \in \R$ as follows
\begin{align*}
B_3 := \frac{1}{ 1 - |v_1^\top \wt{E}_{u_t}| / (\lambda_1 |v_1^\top u_t|^{p-1} )}
\end{align*}
and
\begin{align*}
B_4 := \frac{|v_j^\top \wt{E}_{u_t}|}{ \lambda_1 |v_1^\top u_t|^{p-1}}
\end{align*}
\end{definition}

\begin{lemma}[Part 3 of Theorem~\ref{thm:general_p_bound_EIu}]
\begin{itemize}
    \item Let everything be defined as in Theorem~\ref{thm:general_p_bound_EIu}. 
    \item Suppose that all of the assumptions in Theorem~\ref{thm:general_p_bound_EIu} hold.
\end{itemize}    
    
Then, Eq.~\eqref{eq:part_3_of_thm} hold.
\end{lemma}

\begin{proof}
{\bf Proof of Part 3.} 

Just like Eq.~\eqref{eq:v1ut1_vjut1} and Part 2, $\frac{|v_j^\top u_{t+1}|}{ |v_1^\top u_{t+1} |}$ can also be upper bounded,
\begin{align*}
\frac{|v_j^\top u_{t+1}|}{ |v_1^\top u_{t+1} | } \leq &~ \frac{\lambda_j |v_j^\top u_t|^{p-1} + |v_j^\top \wt{E}_{u_t}|}{ \lambda_1|v_1^\top u_t|^{p-1} - |v_1^\top  \wt{E}_{u_t}|} \\
=  &~ \frac{|v_j^\top u_t|}{ |v_1^\top u_t|}  \cdot \frac{\lambda_j |v_j^\top u_t|^{p-2} + |v_j^\top \wt{E}_{u_t} |/ |v_j^\top u_t|}{ \lambda_1|v_1^\top u_t|^{p-2} - |v_1^\top \wt{E}_{u_t} | / |v_1^\top u_t|} \\
=  &~ \frac{|v_j^\top u_t|}{ |v_1^\top u_t|}  \cdot \frac{\lambda_j |v_j^\top u_t|^{p-2} }{ \lambda_1|v_1^\top u_t|^{p-2} - |v_1^\top \wt{E}_{u_t} | / |v_1^\top u_t|} + \frac{|v_j^\top \wt{E}_{u_t}|}{ \lambda_1 |v_1^\top u_t|^{p-1} - |v_1^\top \wt{E}_{u_t} | } \\
=  &~ \frac{|v_j^\top u_t|}{ |v_1^\top u_t|}  \cdot \frac{\lambda_j |v_j^\top u_t|^{p-2} }{ \lambda_1|v_1^\top u_t|^{p-2} - |v_1^\top \wt{E}_{u_t} | / |v_1^\top u_t|} + \frac{1}{ 1 - |v_1^\top \wt{E}_{u_t}| / (\lambda_1 |v_1^\top u_t|^{p-1} )}  \cdot  \frac{|v_j^\top \wt{E}_{u_t}|}{ \lambda_1 |v_1^\top u_t|^{p-1}}\\
\leq  &~ \frac{|v_j^\top u_t|}{ |v_1^\top u_t|} \cdot \frac{1}{4} \cdot \frac{\lambda_1 |v_1^\top u_t|^{p-2} }{ \lambda_1|v_1^\top u_t|^{p-2} - |v_1^\top \wt{E}_{u_t} | / |v_1^\top u_t|} + \frac{1}{ 1 - |v_1^\top \wt{E}_{u_t}| / (\lambda_1 |v_1^\top u_t|^{p-1} )}  \cdot  \frac{|v_j^\top \wt{E}_{u_t}|}{ \lambda_1 |v_1^\top u_t|^{p-1}}\\
= & ~ \frac{|v_j^\top u_t|}{ |v_1^\top u_t|}  \cdot \frac{1}{4} \cdot \underbrace{ \frac{1}{ 1 - |v_1^\top \wt{E}_{u_t}| / (\lambda_1 |v_1^\top u_t|^{p-1} )} }_{B_3}+ \underbrace{ \frac{1}{ 1 - |v_1^\top \wt{E}_{u_t}| / (\lambda_1 |v_1^\top u_t|^{p-1} )} }_{B_3} \cdot \underbrace{ \frac{|v_j^\top \wt{E}_{u_t}|}{ \lambda_1 |v_1^\top u_t|^{p-1}} }_{B_4},
\end{align*}
where the first step follows from the relationship between the first step and the third step of Eq.~\eqref{eq:v1ut1_vjut1}, the second step follows from simple algebra, the third step follows from simple algebra, the fourth step follows from simple algebra, and the fifth step follows from Part 1 $\lambda_j |v_j^\top u_t|^{p-2} \leq \frac{1}{4} \lambda_1 |v_1^\top u_t|^{p-2}$, and the last step follows from simple algebra.

Similar to Part 1, we can show 
$B_3 \leq 1.1$ 
if 
\begin{align*}
    |v_1^\top u_t| \geq 1 - \frac{1}{c_0^2 p^2 k^2}.
\end{align*}

Then, we consider the bound for $B_4$.  

There are two different situations, namely,
\begin{itemize}
    \item Case 1. $|v_1^\top u_t| \leq 1-\frac{1}{c_0^2 p^2 k^2}$
    \item Case 2. $|v_1^\top u_t|> 1-\frac{1}{c_0^2 p^2 k^2}$.
\end{itemize}

If 
\begin{align*}
    |v_1^\top u_t| \leq (1-\frac{1}{c_0^2 p^2 k^2}),
\end{align*}
we have
\begin{align*}
B_4 = & ~ \frac{| v_j^\top \wt{E}_{u_t}| }{ \lambda_1 | v_1^\top u_t|^{p-1} }\\
= & ~ \frac{\sqrt{1-|v_1^\top u_t|^2 }}{ |v_1^\top u_t|} \cdot \frac{| v_j^\top \wt{E}_{u_t}| }{ \lambda_1 | v_1^\top u_t|^{p-2} \sqrt{1-|v_1^\top u_t|^2} } \\
= & ~ \tan\theta( v_1, u_t) \cdot \frac{ | v_j^\top \wt{E}_{u_t}| }{ \lambda_1 | v_1^\top u_t|^{p-2} \sqrt{1-|v_1^\top u_t|^2} } \\
\leq & ~ \tan\theta( v_1, u_t) \cdot \frac{ c_0pk | v_j^\top \wt{E}_{u_t}| }{ \lambda_1 | v_1^\top u_t|^{p-2}},
\end{align*}
where the 1st step is from the definition of $B_4$, the 2nd step comes from simple algebra, the 3rd step is due to the definition of $\tan\theta( v_1, u_t)$, and the last step follows from $1/\sqrt{1-|v_1^\top u_t|^2} \leq c_0pk$.

We want to find the bound for
\begin{align*}
    \frac{ c_0pk \| \wt{E}_{u_t}\|_2 }{ \lambda_1 | v_1^\top u_t|^{p-2} }.
\end{align*}

We can get that 
\begin{align*}
   \lambda_1 |v_1^\top u_t|^{p-2}  \geq \lambda_1 /(n^{(p-2)/2}  ).
 \end{align*} 
Additionally, based on Part 2 of Lemma~\ref{lem:kappa_phi} and what we assumed about $\ov{E}$ and $E$,
 \begin{align*}
c_0 p k  | v_j^\top \wt{E}_{u_t} | \leq c_0 p k \cdot 4 \epsilon /\sqrt{n}.
 \end{align*}
 Therefore, whenever there is a small enough $\epsilon$ satisfying 
 \begin{align*}
     \epsilon \leq \lambda_1 \sqrt{n} /( n^{(p-2)/2} \cdot 40 \cdot c_0 p k),
 \end{align*}
 then 
 \begin{align*}
     B_4 \leq 0.1 \tan\theta(v_1,u_t).
 \end{align*}

If 
\begin{align*}
    |v_1^\top u_t|> 1-\frac{1}{c_0^2 k^2 p^2},
\end{align*}
then we have
\begin{align*}
B_4 = & ~ \frac{ | v_j^\top \wt{E}_{u_t}| }{ \lambda_1 | v_1^\top u_t|^{p-1} }\\
\leq & ~ \frac{ | v_j^\top \wt{E}_{u_t}| }{ \lambda_1 (1-\frac{1}{c_0^2p^2 k^2})^{p-1} }\\
\leq & ~ 3 | v_j^\top \wt{E}_{u_t}| / \lambda_1, 
\end{align*}
where the first step follows from the definition of $B_4$, the second step follows from $|v_1^\top u_t|> 1-\frac{1}{c_0^2p^2 k^2}$, the third step follows from $\forall p\geq 3,k\geq 1, c_0\geq 1$, we have 
\begin{align*}
    1/(1-\frac{1}{c_0^2p^2k^2})^{p-1}\leq 3.
\end{align*}

By Part 1 of Corollary~\ref{cor:kappa_phi}, we have 
\begin{align*}
    |v_j^\top \wh{E}_{u_t} | \leq 4\epsilon /(c_0\sqrt{n}).
\end{align*}

Based on what we have assumed about $E$ and $\ov{E}$,  
\begin{align*}
    | v_j^\top {E}_{u_t} | \leq \epsilon/(c_0\sqrt{n})
\end{align*}
and 
\begin{align*}
    | v_j^\top \ov{E}_{u_t} | \leq \epsilon /(c_0\sqrt{n}).
\end{align*}

Therefore, the proof of $B_4 \leq 18 \epsilon /(c_0\lambda_1 \sqrt{n})$ is completed.
\end{proof}

\subsection{\texorpdfstring{$\epsilon$}{}-close}

In this section, we upper bound some terms.

\label{sub:more_analysis:epsilon}

\begin{definition}\label{def:epsilon_close}
For any $ \epsilon >0$, we say $\{\wh{\lambda}_i, \wh{v}_i\}_{i=1}^k$ is $\epsilon$-close to $\{ \lambda_i,v_i\}_{i=1}^k$ if for all $i\in [k]$,
\begin{enumerate}
\item $|\wh{\lambda}_i - \lambda_i |\leq \epsilon$.
\item $ \| \wh{v}_i -v_i \|_2\leq \tan \theta(\wh{v}_i,v_i) \leq \min (\sqrt{2}, {\epsilon}/(\lambda_i) )$.
\item $|\wh{v}_i^\top v_j| \leq \epsilon / ( \sqrt{n} \lambda_i)$, $\forall j\in [k]\backslash [i]$.
\end{enumerate}
\end{definition}

\begin{definition}[$A_i$ and $B_i$]\label{def:Ai_Bi}
    We define
    \begin{align*}
        A_i := \lambda_i a_i^{p-1} v_i - \wh{\lambda}_i(a_ic_i+ \| \wh{v}_i^{\bot}\|_2 b_i)^{p-1} c_i 
    \end{align*}
    and
    \begin{align*}
        B_i := \wh{\lambda}_i (a_i c_i + \| \wh{v}_i^{\bot} \|_2 b_i )^{p-1}\notag \| \wh{v}_i^{\bot} \|_2.
    \end{align*}
\end{definition}

\begin{assumption}\label{asm:epsilon}
    We assume that $\epsilon$ is a real number that satisfies
    \begin{align*}
        \epsilon < 10^{-5}\frac{\lambda_k}{p^2 k}.
    \end{align*}
\end{assumption}

\begin{lemma}\label{lem:kappa_phi}
If the following conditions hold
\begin{itemize}
    \item For all $i\in[k]$, $\wh{E}_i = \lambda_i v_i^{\otimes p} - \wh{\lambda}_i \wh{v}_i^{\otimes p}\in \mathbb{R}^{n^p}$.
    \item Let $\epsilon>0$.
    \item $\{\wh{\lambda}_i, \wh{v}_i\}_{i=1}^k$ is $\epsilon$-close to $\{ \lambda_i,v_i\}_{i=1}^k$.
    \item Let $r\in [k]$.
    \item Let $u\in \mathbb{R}^n$ be an unit vector.
\end{itemize}

Then, we have
\begin{enumerate}
\item $  \left\| \overset{r}{\underset{i=1}{\sum}} \wh{E}_i (I,u,\cdots,u) \right\|_2 \leq 2 p\epsilon \kappa^{1/2} + 2\phi \epsilon$.
\item For all $[k] \backslash [i]$, $\left| \overset{r}{\underset{i=1}{\sum}} \wh{E}_j (v_j,u,\cdots,u) \right| \leq  ( 2\kappa \epsilon + \phi \epsilon )/\sqrt{n}$.
\end{enumerate}
where 
\begin{align}\label{eq:kappa}
    \kappa=2\sum_{i=1}^r |u^\top v_i|^2
\end{align}
and
\begin{align}\label{eq:phi}
    \phi = 2k (\epsilon/\lambda_k)^{p-1}.
\end{align}

\end{lemma}
\begin{proof}

{\bf Proof of Part 1.}

Let $i$ be an arbitrary element in $[r]$.

We have that $\wh{E}_i$ is the error and it satisfies
\begin{align}\label{eq:error_wh_E}
\wh{E}_i(I,u,\cdots,u) = \lambda_i (u^\top v_i)^{p-1} v_i - \wh{\lambda}_i (u^\top \wh{v}_i)^{p-1} \wh{v}_i,
\end{align}
which is in the span of $\{v_i, \wh{v}_i\}$. 

Also, the span of $\{v_i, \wh{v}_i\}$ is identical to the span of $\{v_i, \wh{v}_i^{\bot}\}$, where
\begin{align*}
\wh{v}_i^{\bot} = \wh{v}_i - (v_i^\top \wh{v}_i) v_i.
\end{align*}
is the projection of $\wh{v}_i$ onto the subspace orthogonal to $v_i$.

Note
\begin{align}\label{eq:vi_v_2_v}
    \| \wh{v}_i - v\|_2^2  =2 (1- v_i^\top \wh{v}_i).
\end{align}

For convenient, we define
\begin{align*}
c_i = \langle v_i, \wh{v}_i \rangle.
\end{align*}

We can rewrite $c_i$ as follows
\begin{align}\label{eq:c_i}
c_i
= & ~ v_i^\top \wh{v}_i \notag\\
= & ~ 1 - \| \wh{v}_i - v_i \|_2^2 /2 \notag\\
\geq & ~ 0,
\end{align}
the first step follows from definition of $c_i$, the second step follows from Eq.~\eqref{eq:vi_v_2_v}, and the last step is because of the assumption that $\| \wh{v}_i - v_i \|_2 \leq \sqrt{2}$, and it implies that $0 \leq c_i \leq 1$. 

We can also get
\begin{align}\label{eq:vi_leq_vi-vi}
\| \wh{v}_i^{\bot} \|_2^2
= & ~ 1-c_i^2 \notag\\
\leq & ~ \| \wh{v}_i - v_i \|_2^2,
\end{align}
which follows from Eq.~\eqref{eq:c_i} and the Pythagorean theorem.

For all $p$ being greater than or equal to $3$, the following bound can be obtained:
\begin{align*}
| 1 - c_i^p  |
= & ~ | 1 - ( 1- \| \wh{v}_i - v_i \|_2^2 /2 )^p |\\
\leq & ~ \frac{p}{2} \| \wh{v}_i - v_i \|_2^2,
\end{align*}
where the 1st step is due to Eq.~\eqref{eq:c_i} and the 2nd step is supported by Fact~\ref{fac:x_p}.

We present the definition of $a_i \in \mathbb{R}^n$ and $b_i \in \mathbb{R}^n$: 
\begin{align*}
    a_i = u^\top v_i
\end{align*}
and
\begin{align*}
    b_i = u^\top ( \wh{v}_i^{\bot} / \| \wh{v}_i^{\bot} \|_2 ).
\end{align*}

$\wh{E}_i(I,u,\cdots,u)$ can be expressed by the coordinate system of $\wh{v}_i^{\bot}$ and $v_i$:
\begin{align}\label{eq:hat_E_i_value}
 & ~\wh{E}_i(I,u,\cdots,u) \notag\\
 = & ~ \lambda_i (u^\top v_i)^{p-1} v_i - \wh{\lambda}_i (u^\top \wh{v}_i)^{p-1} \wh{v}_i \notag\\
 = & ~ \lambda_i a_i^{p-1} v_i - \wh{\lambda}_i (a_i c_i + \| \wh{v}_i^{\bot} b_i \|_2)^{p-1} (c_i v_i +\wh{v}_i^{\bot})\notag \\
 = & ~ \underbrace{ ( \lambda_i a_i^{p-1} v_i - \wh{\lambda}_i(a_ic_i+ \| \wh{v}_i^{\bot}\|_2 b_i)^{p-1} c_i ) }_{A_i} \cdot v_i - \underbrace{ \wh{\lambda}_i (a_i c_i + \| \wh{v}_i^{\bot} \|_2 b_i )^{p-1}\notag \| \wh{v}_i^{\bot} \|_2 }_{B_i} \cdot \wh{v}_i^{\bot}/\| \wh{v}_i^{\bot} \|_2 \notag\\
 = & A_i \cdot v_i -  B_i \cdot (\wh{v}_i^{\bot} / \| \wh{v}_i^{\bot}\|_2),
\end{align}
where the first step follows from Eq.~\eqref{eq:error_wh_E}, the second step follows from the definition of $a_i$ and $b_i$, the third step follows from simple algebra, and the last step follows from the definition of $A_i$ and $B_i$ (see Definition~\ref{def:Ai_Bi}).

We can express the overall error by:
\begin{align}\label{eq:overall_error}
\left\| \sum_{i=1}^r \wh{E}_i (I,u,\cdots,u) \right\|_2^2
= & ~ \left\| \sum_{i=1}^r A_i v_i - \sum_{i=1}^t B_i (\wh{v}_i^{\bot} / \|\wh{v}_i^{\bot} \|_2) \right\|_2^2 \notag\\
\leq & ~ 2 \left\|\sum_{i=1}^r A_i v_i \right\|_2^2 + 2 \left\| \sum_{i=1}^r B_i (\wh{v}_i^{\bot}/\|  \wh{v}_i^{\bot} \|_2) \right\|_2^2 \notag\\
\leq & ~ 2 \sum_{i=1}^r A_i^2 + 2 \left( \sum_{i=1}^r |B_i| \right)^2,
\end{align}
where the first step follows from Eq.~\eqref{eq:hat_E_i_value},
the second step follows from triangle inequality,
and the third step comes from the definition of the $\ell_2$ norm.

We have
\begin{align}\label{eq:vi_leq_epsilon}
    \|\wh{v}_i^{\bot}\|_2^2 
    \leq & ~ \| \wh{v}_i -v_i\|_2^2 \notag\\
    \leq & ~ \epsilon / \lambda_i,
\end{align}
where the first step follows from Eq.~\eqref{eq:vi_leq_vi-vi} and the second step follows from Definition~\ref{def:epsilon_close}.

By using 
\begin{align}\label{eq:lambda_lambda_leq_epsilon}
    |\lambda_i -\wh{\lambda}_i| \leq \epsilon,
\end{align}
we first try to bound $A_i$ for $|b_i|$ being smaller than $1$ and $c_i \in [0, 1]$,
\begin{align}\label{eq:bound_Ai}
|A_i|
= & ~ | \lambda_i a_i^{p-1}   - \wh{\lambda}_i (a_i c_i + \| \wh{v}_i^{\bot}\|_2 b_i )^{p-1} c_i | \notag\\
\leq & ~ | \lambda_i a_i^{p-1} - \wh{\lambda}_i c_i^p a_i^{p-1}| + \sum_{j=1}^{p-1} \wh{\lambda}_i c_i {(p-1) \choose j} | a_i c_i|^{ (p-1) -j } \| \wh{v}_i^{\bot}\|_2^j \notag\\
= & ~ | \lambda_i a_i^{p-1}   - \wh{\lambda}_i c_i^p a_i^{p-1} + \wh{\lambda}_i a_i^{p-1} - \wh{\lambda}_i   a_i^{p-1}| + \sum_{j=1}^{p-1} \wh{\lambda}_i c_i {(p-1) \choose j} | a_i c_i|^{ (p-1) -j } \| \wh{v}_i^{\bot}\|_2^j \notag\\
\leq & ~ | \lambda_i a_i^{p-1} - \wh{\lambda}_i   a_i^{p-1}| + |\wh{\lambda}_i a_i^{p-1}  - \wh{\lambda}_i c_i^p a_i^{p-1} | + \sum_{j=1}^{p-1} \wh{\lambda}_i c_i {(p-1) \choose j} | a_i c_i|^{ (p-1) -j } \| \wh{v}_i^{\bot}\|_2^j \notag\\
\leq & ~ | \lambda_i a_i^{p-1} - \wh{\lambda}_i   a_i^{p-1}| + |(1-c_i^p)\wh{\lambda}_i   a_i^{p-1} | + \sum_{j=1}^{p-1} \wh{\lambda}_i c_i {(p-1) \choose j} | a_i c_i|^{ (p-1) -j } \| \wh{v}_i^{\bot}\|_2^j \notag\\
\leq & ~ | \lambda_i a_i^{p-1}   - \wh{\lambda}_i   a_i^{p-1}| + | (1-c_i^p)\wh{\lambda}_i   a_i^{p-1}|+ \sum_{j=1}^{p-1} \wh{\lambda}_i c_i {(p-1) \choose j} | a_i c_i|^{ (p-1) -j } (\epsilon /\lambda_i)^{j} \notag\\
\leq & ~ \epsilon |a_i|^{p-1} + \frac{p}{2} (\epsilon/\lambda_i)^2 \wh{\lambda}_i |a_i|^{p-1} + \sum_{j=1}^{p-1} \wh{\lambda}_i {(p-1) \choose j} | a_i |^{ (p-1) -j } (\epsilon /\lambda_i)^{j},
\end{align}
where the first step follows from the definition of $A_i$ (see Definition~\ref{def:Ai_Bi}), the second step follows from  binomial theorem and $|b_i| < 1$, the third step follows from adding and subtracting the same thing, the fourth step follows from triangle inequality, the fifth step follows from simple algebra, the sixth step follows from Eq.~\eqref{eq:vi_leq_epsilon}, and the last step follows from Eq.~\eqref{eq:lambda_lambda_leq_epsilon}.

Note that the second term of Eq.~\eqref{eq:bound_Ai} can be bounded as
\begin{align}\label{eq:bound_epsilon/lambda}
\frac{p}{2} (\epsilon/\lambda_i)^2 \wh{\lambda}_i |a_i|^{p-1} 
\leq & ~ \frac{p}{2} (\epsilon/\lambda_i)^2 ( |\wh{\lambda}_i-\lambda_i| + |\lambda_i| ) |a_i|^{p-1} \notag\\
= & ~ \frac{p}{2} (\epsilon/\lambda_i)^2 |\wh{\lambda}_i-\lambda_i| |a_i|^{p-1} + \frac{p}{2} (\epsilon/\lambda_i)^2 |\lambda_i| |a_i|^{p-1} \notag\\
\leq & ~ \frac{p}{2} (\epsilon/\lambda_i)^2 \epsilon |a_i|^{p-1} + \frac{p}{2} (\epsilon/\lambda_i)^2 |\lambda_i| |a_i|^{p-1}\notag\\
\leq & ~ \frac{p}{2} (\epsilon/\lambda_i)^2 \epsilon |a_i|^{p-1} + \frac{p}{2} (\epsilon/\lambda_i) \cdot \epsilon |a_i|^{p-1} \notag\\
\leq & ~ \frac{1}{100 k} \epsilon |a_i|^{p-1},
\end{align}
where the first step follows from triangle inequality, the second step follows from simple algebra, the third step follows from Eq.~\eqref{eq:lambda_lambda_leq_epsilon}, the fourth step follows from Eq.~\eqref{eq:vi_leq_epsilon}, and the last step follows from Assumption~\ref{asm:epsilon}.

We can separate the third term of Eq.~\eqref{eq:bound_Ai} into two components
\begin{enumerate}
    \item $j \in \{1, \cdots, (p-1)/2\}$ and
    \item $j \in \{(p-1)/2,\cdots, (p-1)\}$.
\end{enumerate}

Consider the first component:
\begin{align*}
\sum_{j=1}^{ (p-1)/2} \wh{\lambda}_i {(p-1) \choose j} | a_i |^{ (p-1) -j } (\epsilon /\lambda_i)^{j} 
=  & ~ \wh{\lambda}_i (p-1) |a_i|^{p-2} \epsilon /\lambda_i +  \sum_{j=2}^{ (p-1)/2} \wh{\lambda}_i (p-1)^j | a_i |^{ (p-1)/2  } (\epsilon /\lambda_i)^{j} \\
\leq & ~ 2(p-1)\epsilon |a_i|^{p-2} + \sum_{j=2}^{ (p-1)/2} \wh{\lambda}_i (p-1)^j | a_i |^{ (p-1)/2  } (\epsilon /\lambda_i)^{j} \\
= & ~ 2(p-1)\epsilon |a_i|^{p-2} + \wh{\lambda}_i | a_i |^{ (p-1)/2 } \sum_{j=2}^{ (p-1)/2} (p-1)^j (\epsilon /\lambda_i)^{j} \\
\leq & ~ 2(p-1)\epsilon |a_i|^{p-2} + \wh{\lambda}_i | a_i |^{ (p-1)/2 } \sum_{j=0}^\infty \left(\frac{1}{2}\right)^j \\
\leq & ~ 2(p-1)\epsilon |a_i|^{p-2} + \wh{\lambda}_i \epsilon \frac{\epsilon (p-1)^2}{\lambda_i^2} \cdot 2 \cdot |a_i|^{(p-1)/2} \\
\leq & ~ 2(p-1)\epsilon |a_i|^{p-2} + \frac{1}{100k} \epsilon |a_i|^{(p-1)/2},
\end{align*}
where the first step is by expanding the summation term, the second step is because of the fact that $\wh{\lambda}_i \leq 2\lambda_i$, the third step is supported by $\sum_i c a_i = c\sum_i a_i$, the fourth step follows from the fact that each term of $\sum_{j=2}^{ (p-1)/2} (p-1)^j (\epsilon /\lambda_i)^{j}$ is bounded by the corresponding term of $\sum_{j=0}^\infty \left(\frac{1}{2}\right)^j$, the fifth step follows from Fact~\ref{fac:geometric_series}, the sixth step follows from Assumption~\ref{asm:epsilon} and $\wh{\lambda}_i \leq 2 \lambda_i$.  

Similarly, we can bound the second component,
\begin{align*}
 & ~ \sum_{j=(p-1)}^{ (p-1)/2} \wh{\lambda}_i {(p-1) \choose j} | a_i |^{ (p-1) -j } (\epsilon /\lambda_i)^{j} \\
 = & ~ \wh{\lambda}_i \cdot 1 \cdot 1 \cdot (\epsilon /\lambda_i)^{p-1}+ \sum_{j=p}^{ (p-1)/2} \wh{\lambda}_i {(p-1) \choose j} | a_i |^{ (p-1) -j } (\epsilon /\lambda_i)^{j} \\
  = & ~ 2 \epsilon^{p-1} /\lambda_i^{p-2}+ \sum_{j=p}^{ (p-1)/2} \wh{\lambda}_i {(p-1) \choose j} | a_i |^{ (p-1) -j } (\epsilon /\lambda_i)^{j} \\
  \leq & ~ 2 \epsilon^{p-1} /\lambda_i^{p-2}+ {(p-1) \choose (p - 1)/2} \sum_{j=p}^{ (p-1)/2} \wh{\lambda}_i | a_i |^{ (p-1) -j } (\epsilon /\lambda_i)^{j} \\
  \leq & ~ 2 \epsilon^{p-1} /\lambda_i^{p-2}+ {(p-1) \choose (p - 1)/2} \wh{\lambda}_i | a_i |^{ (p-1)/2 } \sum_{j=p}^{ (p-1)/2} (\epsilon /\lambda_i)^{j} \\
  \leq & ~ \frac{1}{100 k} \epsilon + {(p-1) \choose (p - 1)/2} \wh{\lambda}_i | a_i |^{ (p-1)/2 } \sum_{j=p}^{ (p-1)/2} (\epsilon /\lambda_i)^{j} \\
  \leq & ~ \frac{1}{100 k} \epsilon + {(p-1) \choose (p - 1)/2} \wh{\lambda}_i | a_i |^{ (p-1)/2 } (\epsilon /\lambda_i)^{(p-1)/2} \sum_{j=p}^{ (p-1)/2} (\epsilon /\lambda_i)^{j - (p-1)/2} \\
  \leq & ~ \frac{1}{100 k} \epsilon + 2 {(p-1) \choose (p - 1)/2} \wh{\lambda}_i | a_i |^{ (p-1)/2 } (\epsilon /\lambda_i)^{(p-1)/2}  \\
  \leq & ~ \frac{1}{100 k} \epsilon + 4 {(p-1) \choose (p - 1)/2} \lambda_i | a_i |^{ (p-1)/2 } (\epsilon /\lambda_i)^{(p-1)/2}  \\
 \leq & ~\frac{1}{100 k} \epsilon + \frac{1}{100 k} \epsilon |a_i|
\end{align*}
where the first step comes from expanding the summation term, the second step can be gotten from $\wh{\lambda}_i \leq 2\lambda_i$, the third step can be supported by $\max_{j} \{{(p-1) \choose j}\} = {(p-1) \choose (p - 1)/2}$, the fourth step follows from $\sum_i c a_i = c\sum_i a_i$, the fifth step follows from Assumption~\ref{asm:epsilon}, the sixth step follows from simple algebra, the seventh step follows from the Fact~\ref{fac:geo_seires_apply}, the eighth step follows from $\wh{\lambda}_i \leq 2 \lambda_i$, and the last step follows from Assumption~\ref{asm:epsilon}.

Thus, putting it all together, we get
\begin{align*}
|A_i| \leq \epsilon |a_i| + \frac{1}{10k} \epsilon |a_i| + (p-1) \epsilon |a_i| + \frac{1}{100k} \epsilon
\end{align*}
which implies that
\begin{align}\label{eq:final_bound_Ai}
|A_i|^2 \leq 2 \left(  (\epsilon |a_i|)^2 + ( \frac{1}{10k} \epsilon |a_i|)^2 + ( (p-1) \epsilon |a_i| )^2+  (\frac{1}{100k} \epsilon)^2 \right)
\end{align}

Next, we need to find the bound of $B_i$,
\begin{align}\label{eq:bound_Bi}
|B_i| 
= & ~ | \wh{\lambda}_i \|\wh{v}_i^{\bot}\|_2 (a_i c_i + \| \wh{v}_i^{\bot} \|_2 b_i)^{p-1}  | \notag\\
\leq & ~ \wh{\lambda}_i \| \wh{v}_i^{\bot} \|_2 \sum_{j=0}^{p-1} {(p-1) \choose j} |a_ic_i|^{p-1 - j} \| \wh{v}_i^{\bot} \|_2^j \notag\\
\leq & ~ \wh{\lambda}_i (\epsilon /\lambda_i) \sum_{j=0}^{p-1} {(p-1) \choose j} |a_i|^{p-1-j} (\epsilon / \lambda_i )^j \notag\\
= & ~ \wh{\lambda}_i (\epsilon /\lambda_i) |a_i|^{p-1} + \wh{\lambda}_i (\epsilon /\lambda_i) \sum_{j=1}^{p-1} {(p-1) \choose j} |a_i|^{p-1-j} (\epsilon / \lambda_i )^j,
\end{align}
where the first step follows from the definition of $B_i$ (see Definition~\ref{def:Ai_Bi}), the second step follows from binomial theorem and $|b_i| < 1$, the third step follows from $\| \wh{v}_i^{\bot} \|_2 \leq \epsilon/\lambda_i$, and the last step follows from extracting the first term from the summation.  

Note that the first term of Eq.~\eqref{eq:bound_Bi} is 
\begin{align*}
    \wh{\lambda}_i (\epsilon /\lambda_i) |a_i|^{p-1},
\end{align*}
which can be bounded as
\begin{align*}
 \wh{\lambda}_i (\epsilon /\lambda_i) |a_i|^{p-1}
 \leq & ~ (\epsilon + \lambda_i) (\epsilon /\lambda_i ) |a_i|^{p-1}\\
 = & ~ \epsilon |a_i|^{p-1} + (\epsilon^2 /\lambda_i ) |a_i|^{p-1}\\
 \leq & ~ \epsilon |a_i|^{p-1} + \frac{1}{100k} \epsilon |a_i|^{p-1}
\end{align*}
where the first step follows from simple algebra, and the second step follows from Eq.~\eqref{eq:bound_epsilon/lambda}.

The second term of Eq.~\eqref{eq:bound_Bi} is 
\begin{align*}
    \wh{\lambda}_i (\epsilon /\lambda_i) \sum_{j=1}^{p-1} {(p-1) \choose j} |a_i|^{p-1-j} (\epsilon / \lambda_i )^j,
\end{align*}

We can separate this into two components
\begin{enumerate}
    \item $j \in \{1, \cdots, (p-1)/2\}$ and
    \item $j \in \{(p-1)/2,\cdots, (p-1)\}$.
\end{enumerate}

The first component is:
\begin{align*}
 & ~ (\epsilon/\lambda_i)\sum_{j=1}^{ (p-1)/2} \wh{\lambda}_i  {(p-1) \choose j} | a_i |^{ (p-1) -j } (\epsilon /\lambda_i)^{j} \\
=  & ~ (\epsilon/\lambda_i) \wh{\lambda}_i (p-1) |a_i|^{p-2} \epsilon /\lambda_i +  (\epsilon/\lambda_i) \sum_{j=2}^{ (p-1)/2} \wh{\lambda}_i (p-1)^j | a_i |^{ (p-1)/2  } (\epsilon /\lambda_i)^{j} \\
\leq & ~ \frac{1}{100k} \epsilon |a_i|^{p-2} + \frac{1}{100k} \epsilon |a_i|^{(p-1)/2}
\end{align*}
where the first step comes from expanding the summation term and the second step is by Assumption~\ref{asm:epsilon}.

The second component is: 
\begin{align*}
 & ~ (\epsilon/\lambda_i) \sum_{j=(p-1)}^{ (p-1)/2} \wh{\lambda}_i {(p-1) \choose j} | a_i |^{ (p-1) -j } (\epsilon /\lambda_i)^{j} \\
 = & ~ (\epsilon/\lambda_i) \wh{\lambda}_i \cdot 1 \cdot 1 \cdot (\epsilon /\lambda_i)^{p-1}+ (\epsilon/\lambda_i) \sum_{j=p}^{ (p-1)/2} \wh{\lambda}_i {(p-1) \choose j} | a_i |^{ (p-1) -j } (\epsilon /\lambda_i)^{j} \\
 \leq & ~\frac{\phi}{ k} \epsilon + \frac{1}{100 k} \epsilon |a_i|
\end{align*}
where the first step follows from expanding the summation term, and the second step follows from $\phi = 2 k (\epsilon/\lambda_k)^{p-1}$ and Assumption~\ref{asm:epsilon}. 

Putting it all together, we have

\begin{align*}
  |B_i |\leq \epsilon |a_i|^{2} + \frac{1}{100k^2} \epsilon |a_i| + \frac{\phi}{k} \epsilon 
 \end{align*}
Taking the summation over all the $r$ terms on both sides, we obtain
 \begin{align*}
 ~\sum_{i=1}^r |B_i| \leq \epsilon \kappa + \frac{1}{100k^2} \epsilon \sum_{i=1}^t |a_i| + \phi \epsilon.
 \end{align*}

Using 
\begin{align*}
    (x+y+z)^2 \leq 3 (x^2 + y^2 + z^2),
\end{align*}
we have
 \begin{align}\label{eq:bound_sum_Bi}
 ( \sum_{i=1}^r |B_i| )^2 \leq & ~ 3\left( (\epsilon \kappa)^2 + (\frac{1}{100k} \epsilon \sum_{i=1}^t |a_i|)^2 + (\phi\epsilon)^2 \right) \notag\\
 \leq & ~  3\left( (\epsilon \kappa)^2 + (\frac{1}{100k} \epsilon )^2 \kappa k + (\phi\epsilon)^2 \right),
\end{align}
where the last step follows from $(\sum_{i=1}^t |a_i|)^2 \leq \kappa k$.

Recall that 
\begin{align*}
    \kappa = \sum_{i=1}^t |a_i|^2 \leq 1.
\end{align*}

In general, we can get
\begin{align*}
 \left\| \sum_{i=1}^r \wh{E}_i(I,u,\cdots,u) \right\|_2^2
 = & ~2 \sum_{i=1}^r |A_i|^2 + 2 (\sum_{i=1}^r |B_i|)^2 \\
 \leq & ~ 4 \left( \epsilon^2 \cdot \kappa + (\frac{1}{10k}\epsilon)^2\kappa + (p-1)^2\epsilon^2 \kappa + (\frac{1}{100\sqrt{k}}\epsilon)^2 \right) \\
+ & ~ 2 (\sum_{i=1}^r |B_i|)^2 \\
\leq & ~ 4 \left( \epsilon^2 \cdot \kappa + (\frac{1}{10k}\epsilon)^2\kappa + (p-1)^2\epsilon^2 \kappa + (\frac{1}{100\sqrt{k}}\epsilon)^2 \right) \\
+ & ~ 4\left( (\epsilon \kappa)^2 + (\frac{1}{100k} \epsilon )^2 \kappa k + (\phi\epsilon)^2 \right) \\
\leq & ~ 4p^2 \epsilon^2 \kappa + 4\phi^2 \epsilon^2
\end{align*}
where the first step follows from Eq.~\eqref{eq:overall_error}, the second step follows from Eq.~\eqref{eq:final_bound_Ai}, the third step follows from Eq.~\eqref{eq:bound_sum_Bi}, and the last step follows from simple algebra.

The desired bound is given by this equation.

{\bf Proof of Part 2.}

Let $j$ be an arbitrary element of $[k] \backslash [i]$.

Now, we can get
\begin{align*}
\left| \sum_{i=1}^r \wh{E}_i(v_j,u,\cdots, u) \right| 
 \leq & ~ \sum_{i=1}^r | \wh{E}_i(v_j,u,\cdots, u) | \\
 = & ~ \sum_{i=1}^r \wh{\lambda}_i |\wh{v}_i^\top u|^{p-1} |\wh{v}_i^\top v_j| \\
 \leq & ~ \sum_{i=1}^t \wh{\lambda}_i |\wh{v}_i^\top u|^{p-1} \frac{\epsilon}{ \sqrt{n} \lambda_i },
\end{align*}
for the 1st step, we use the triangle inequality, for the 2nd step, we utilize the fact that $ \langle v_j, v_i \rangle = 0, \forall i\neq j$, and for the last step, we employ the third part of the definition of $\epsilon$-close.

Now, we analyze the bound for $\sum_{i=1}^r \frac{\wh{\lambda}_i}{\lambda_i} |\wh{v}_i^\top u|^{p-1}$:
\begin{align*}
 \sum_{i=1}^r \frac{ \wh{\lambda}_i }{\lambda_i} |\wh{v}_i^\top u|^{p-1}
\leq & ~\sum_{i=1}^r \frac{ \wh{\lambda}_i }{\lambda_i} (|v_i^\top u| + | (v_i-\wh{v}_i)^\top u| )^{p-1} \\
\leq & ~\sum_{i=1}^r \frac{ \wh{\lambda}_i }{\lambda_i} (|v_i^\top u| + \| v_i-\wh{v}_i\|_2 )^{p-1} \\
\leq & ~\sum_{i=1}^r (1 +\epsilon/\lambda_i) \cdot (|v_i^\top u| + \epsilon/\lambda_i )^{p-1} \\
\leq & ~ 2 \sum_{i=1}^r (|v_i^\top u| + \epsilon/\lambda_i )^{p-1}  \\
\leq & ~ 2 \sum_{i=1}^r 2^{p-2} \cdot |v_i^\top u|^{p-1} + 2^{p-2} \cdot ( \epsilon/\lambda_i )^{p-1} \\
\leq & ~ 2\sum_{i=1}^r 2 |v_i^\top u|^2 +   ( 2 \epsilon/\lambda_i )^{p-1}  \\
\leq & ~ 2\sum_{i=1}^r 2 |v_i^\top u|^2 +  2k (2\epsilon/\lambda_k)^{p-1} \\
= & ~ 2\kappa + \phi.
\end{align*}
where the first step follows from the triangle inequality, the second step follows from Cauchy-Scharwz inequality and $\| u\|_2 \leq 1$, the third step follows from Eq.~\eqref{eq:vi_leq_epsilon}, the fourth step follows from $\epsilon/\lambda_i \leq 2$, the fifth step follows from Fact~\ref{fac:x_p}, the sixth step follows from $|v_i^\top u| < 1/4$ and $p \geq 3$, the seventh step follows from $\lambda_k \leq \lambda_i$, and the last step follows from the definition of $\phi$ and $\kappa$ (see Eq.~\eqref{eq:phi} and Eq.~\eqref{eq:kappa}).

Then, we complete the proof with the desired bound $(2\kappa + \phi)\epsilon /\sqrt{n}$.

\end{proof}

\begin{corollary}\label{cor:kappa_phi}
If the following conditions hold:
\begin{itemize}
    \item For all $i\in[k]$, let $\wh{E}_i = \lambda_i v_i^{\otimes p} - \wh{\lambda}_i \wh{v}_i^{\otimes p}\in \mathbb{R}^{n^p}$.
    \item Let $c_0\geq 1$.
    \item Let $r\in [k]$.
    \item Let $\epsilon \leq \lambda_k /(2c_0 k )$.
    \item Suppose that $\{\wh{\lambda}_i, \wh{v}_i\}_{i=1}^k$ is $\epsilon$-close to $\{ \lambda_i,v_i\}_{i=1}^k$.
    \item $u\in \mathbb{R}^n$ is an unit vector.
    \item Suppose $|u^\top v_{r+1}| \geq 1-\frac{1}{c_0^2 p^2 k}$.
    \item Let $\kappa=\overset{r}{\underset{i=1}{\sum}} |u^\top v_i|^2$.
    \item Let $\phi = 2k (\epsilon/\lambda_k)^{p-1}$.
\end{itemize}

Then,
\begin{enumerate}
\item $  \left\| \overset{r}{\underset{i=1}{\sum}} \wh{E}_i (I,u,\cdots,u) \right\|_2 \leq 2 p\epsilon \kappa^{1/2} + 2\phi \epsilon \leq 4\epsilon/c_0$.
\item $ \forall j \in [k] \backslash [i] , \left| \overset{r}{\underset{i=1}{\sum}} \wh{E}_j (v_j,u,\cdots,u) \right| \leq (2\kappa \epsilon + \phi \epsilon)/\sqrt{n} \leq 4\epsilon/(c_0 \sqrt{n})$.
\end{enumerate}

\end{corollary}
\begin{proof}
Based on Fact~\ref{fac:general_p_three_vectors}, we can get that for any arbitrary $i$ in $[r]$,
\begin{align*}
\kappa
= & ~ \sum_{i=1}^r |u^\top v_{i} |^2\\
\leq & ~ k \cdot 1/ (c_0^2 p^2 k)\\
= & ~ 1/c_0^2 p^2,
\end{align*}
where the first step follows from definition of $\kappa$, the second step follows from $r \leq k$ and $\max_i |u^\top v_i|^2 \leq 1/(c_0^2 p^2 k)$, and the last step follows from simple algebra.

This implies 
\begin{align*}
    2p \epsilon \sqrt{\kappa} \leq 2/c_0.
\end{align*}

We can also bound $\phi$,
\begin{align*}
\phi
= & ~ 2k (\epsilon /\lambda_k)^{p-1}\\
\leq & ~ 2k (1/2c_0 k)^{p-1} \\
\leq & ~ 2k (1/ (2 c_0 k) )^2\\
\leq & ~ 1/c_0.
\end{align*}
where the 1st step can be gotten from the definition of $\phi$, the 2nd step is because of $\epsilon \leq \lambda_k / (2 c_0 k)$ (from Corollary statement), the 3rd step is due to $p \geq 2$, and the 4th step can be seen from $2k \geq 1$.

Therefore, we complete our proof.
\end{proof}

%% file: combine.tex
\section{Combine}
\label{sec:combine}

In this section, we present Theorem~\ref{thm:general_p_without_sketch} and Theorem~\ref{thm:main_formal} and prove them.

\begin{theorem}[Arbitrary order robust tensor power method, formal version of Lemma~\ref{lem:informal_version}]\label{thm:general_p_without_sketch}

If the following conditions hold
\begin{itemize}
    \item Let $p$ be greater than or equal to $3$.
    \item Let $k$ be greater than or equal to $1$.
    \item Let $\lambda_i > 0$.
    \item With $n \geq k$, $\{v_1, \ldots, v_k\} \subseteq \mathbb{R}^n$ is an orthonormal basis vectors.
    \item Let $A  = A ^* +E \in \mathbb{R}^{n^p}$ be an arbitrary tensor satisfying $A ^* = \sum_{i=1}^k \lambda_i v_i^{\otimes p}$.
    \item Suppose that $\lambda_{1}$ is the greatest values in $\{\lambda_i\}_{i=1}^k$.
    \item Suppose that $\lambda_{k}$ is the smallest values in $\{\lambda_i\}_{i=1}^k$.
    \item The outputs obtained from the robust tensor power method are $\{\wh{\lambda}_i, \wh{v}_i\}_{i=1}^k$.
    \item $E$ satisfies that $\| E \| \leq \epsilon / (c_0 \sqrt{n} )$ (see definition of $\| E \|$ in Section~\ref{sec:prelim}).
    \item $T = \Omega( \log(\lambda_{1}  n/\epsilon) )$.
    \item $L = \Omega( k \log(k))$.
    \item $c_0\geq100$ and $c>0$
    \item For all $\epsilon$ satisfying $\epsilon \in (0, c \lambda_k / ( c_0  p^2 k n^{(p-2)/2})$.
\end{itemize}

  Then, with probability at least $9/10$, there exists a permutation $\pi:[k] \rightarrow [k]$, such that $\forall i\in[k]$,
  \begin{align}\label{eq:guarantee}
  |\lambda_i - \wh{\lambda}_{\pi(i)}| \leq \epsilon, \ \ \ \
  \|v_i - \wh{v}_{\pi(i)}\|_2 \leq \epsilon/ \lambda_i.
  \end{align}
 \end{theorem}

\begin{proof}
Let $E \in \mathbb{R}^{n^p}$ be the original noise.

Let
\begin{align*}
    \wh{E}_i = \lambda_i v_i^{\otimes p} - \wh{\lambda}_i \wh{v}_i^{\otimes p} \in \mathbb{R}^{n^p}
\end{align*}
be the deflation noise.

$\ov{E} \in \mathbb{R}^{n^p}$ represents the sketch noise.

  $\wt{E}$ represents the ``true" noise, including all the original, deflation and sketch noises. 

As a result, for the $t+1$ step, we analyze $A ^*+\wt{E}$, which is a tensor satisfying 
\begin{align*}
    \wt{E}=E+\sum_{i=1}^t \wh{E}_i + \ov{E}.
\end{align*}

There is no need for us to consider $\ov{E}$, the sketch noise. However, to prove a stronger statement, we do not regard $\ov{E}$ to be equal to $0$, but only assume that it is bounded, namely
\begin{align}\label{eq:bound_ovE}
    \| \ov{E}\| \leq \epsilon /(c_0 \sqrt{n}).
\end{align}

We use mathematical induction to proof this.

{\bf Base case.} 

Let $i = 1$.

For the 1st step, we have that $\wh{\lambda}_1 \in \mathbb{R}$ and $\wh{v}_1 \in \mathbb{R}^n$.

As Part 2 of Definition~\ref{def:epsilon_close}, we show 
\begin{align*}
    \| \wh{v}_1 -  v_1 \|_2 
\end{align*}
is bounded. 

Then, as Part 1 of Definition~\ref{def:epsilon_close}, we show 
\begin{align*}
    |\wh{\lambda}_1 - \lambda_1|
\end{align*}
is bounded. 

At the end, as Part 3 of Definition~\ref{def:epsilon_close}, we show 
\begin{align*}
    |\wh{v}_i^\top v_j|
\end{align*}
is bounded.

{\bf Bounding $|\wh{v}_1 - v_1|$.}

We have
\begin{align}\label{eq:bound_tan_theta}
\tan \theta(u_0, v_1)
= & ~ \sin\theta(u_0,v_1) /\cos\theta(u_0,v_1) \notag\\
= & ~ \sqrt{1 - \langle u_0, v_1 \rangle^2} /\langle u_0, v_1 \rangle \notag\\
= & ~ \sqrt{\frac{1 - \langle u_0, v_1 \rangle^2}{\langle u_0, v_1 \rangle^2}}  \notag\\
= & ~ \sqrt{\frac{1}{\langle u_0, v_1 \rangle^2} - 1}  \notag\\
\leq & ~ \sqrt{\frac{1}{\langle u_0, v_1 \rangle^2}}  \notag\\
= & ~ \frac{1}{\langle u_0, v_1 \rangle} \notag\\
\leq & ~ \sqrt{n},
\end{align}
where the first step follows from Definition~\ref{def:sin_cos_tan}, the second step follows from Definition~\ref{def:sin_cos_tan}, the third step follows from simple algebra, the fourth step follows from simple algebra, the fifth step follows from simple algebra, the sixth step follows from simple algebra, and the last step follows from Lemma~\ref{lem:wang_lem_C1}.

$t^*$ represents the condition for 
 \begin{align}\label{eq:ut*_v1_equal_to}
     |u_{t^*}^\top v_1 | = 1 - \frac{1}{c_0^2 p^2 k^2}.
 \end{align}
 
 We know 
 \begin{align*} 
  \| u_{t^*}- v_1 \|_2^2 
  = & ~ \| u_{t^*}\|_2^2 + \| v_1 \|_2^2 - 2 \langle u_{t^*} , v_1\rangle \\
  = & ~ 1 + 1 - 2 \langle u_{t^*}, v_1 \rangle \\
  = & ~ 2 - 2 |u_{t^*}^\top v_1| \\
  = & ~ 2 - 2 (1 - \frac{1}{c_0^2 p^2 k^2}) \\
  = & ~ 2/(c_0^2 p^2 k^2),
 \end{align*}
 where the first step follows from simple algebra, the second step follows from the fact that $u_{t^*}$ and $v_1$ are unit vectors, the third step is because the inner product is positive, the fourth step follows from Eq.~\eqref{eq:ut*_v1_equal_to}, and the last step follows from simple algebra.
 
 We can upper bound 
\begin{align*}
  \| u_{t^*}- v_1 \|_2 
  \leq & ~ \tan \theta( u_{t^*}, v_1 ) \\
 \leq & ~ 0.8 \tan\theta(u_{t^*-1},v_1) \\
 \leq & ~ \cdots \\
\leq & ~ 0.8^{t^*} \tan\theta(u_0,v_1) \\
\leq & ~ 0.8^{t^*} \sqrt{n},
\end{align*}
where the first step is due to Fact~\ref{fac:bound_u-v}, the second can be seen from Part 1 of Theorem~\ref{thm:general_p_bound_EIu}, the second last step can be gotten from Part 1 of Theorem~\ref{thm:general_p_bound_EIu}, and the last step follows from Eq.~\eqref{eq:bound_tan_theta}.

After that, we let
\begin{align*}
    t^* = \Omega( \log(nkpc_0) ) = \Omega(\log (c_0 n) ).
\end{align*}

For $\| u_T - v_1\|_2$, we can show
\begin{align*}
\| u_T - v_1\|_2 
\leq & ~ 0.8 \tan\theta(u_T,v_1) + 18 \epsilon /(c_0 \lambda_1)  \\
\leq & ~ \cdots \\
\leq & ~ 0.8^{T-t^*} \tan (u_{t^*},v_1) + 5 \cdot 18 \epsilon /(c_0 \lambda_1),
\end{align*}
where the first step follows from Part 1 of Theorem~\ref{thm:general_p_bound_EIu}, and the last step follows from recursively applying Part 1 of Theorem~\ref{thm:general_p_bound_EIu}.

To guarantee 
\begin{align*}
    \| u_T - v_1\|_2 \leq \epsilon / \lambda_1,
\end{align*}
we let
\begin{align*}
    T-t^* = \Omega(n \lambda_1 /\epsilon )
\end{align*}
and $c_0 \geq 100$.

Therefore, we achieve the intended property as outlined in Part 2 of Definition~\ref{def:epsilon_close}.

{\bf Bounding $|\wh{\lambda}_1 - \lambda_1|$.}

It remains to bound $|\wh{\lambda}_1 - \lambda_1 |$.
\begin{align}\label{eq:bound_lambda1-lambda1}
|\wh{\lambda}_1 - \lambda_1 | = & ~ | [A ^* +\wt{E}](\wh{v}_1,\cdots, \wh{v}_1) - \lambda_1 | \notag\\
\leq & ~ |\wt{E}(\wh{v}_1,\cdots, \wh{v}_1)| + | A ^*(\wh{v}_1,\cdots,\wh{v}_1 ) - \lambda_1  | \notag\\
= & ~ |\wt{E}(\wh{v}_1,\cdots, \wh{v}_1)| +  \left| \left[\sum_{i=1}^k \lambda_i v_i^{\otimes p}\right](\wh{v}_1,\cdots,\wh{v}_1 ) - \lambda_1  \right| \notag\\
\leq & ~ \underbrace{ | \wt{E}(\wh{v}_1,\cdots, \wh{v}_1)| }_{B_5} + \underbrace{ |\lambda_1 |v_1^\top \wh{v}_1|^p -\lambda_1| }_{B_6} + \underbrace{ \sum_{j=2}^k \lambda_j |v_j^\top \wh{v}_1|^p }_{B_7}, 
\end{align}
where the first step follows from the definition of $\wh{\lambda}_1$, the second step follows from the triangle inequality, the third step follows from
\begin{align*}
    A ^* = \sum_{i=1}^k \lambda_i v_i^{\otimes p},
\end{align*}
and the last step follows from the triangle inequality.

For the term $B_5$, we have
\begin{align}\label{eq:B5}
B_5 = & ~ | \wt{E} (\wh{v}_1,\cdots,\wh{v}_1)|  \notag\\
\leq & ~ | E(\wh{v}_1,\cdots,\wh{v}_1) | + | \ov{E}(\wh{v}_1, \cdots, \wh{v}_1) | \notag\\ 
\leq & ~ \| E\| + | \ov{E}(\wh{v}_1, \cdots, \wh{v}_1) | \notag\\
\leq & ~ \epsilon/(c_0 \sqrt{n}) + \epsilon / ( c_0 \sqrt{n}) \notag\\
\leq & ~ \epsilon /12,
\end{align}
where the first step follows from the definition of $B_5$ (see Eq.~\eqref{eq:bound_lambda1-lambda1}), the second step follows from triangle inequality, the third step follows from the definition of tensor spectral norm, the fourth step follows from Eq.~\eqref{eq:bound_ovE}, and the last step follows from $c_0 \geq 100$ and $n$ is greater than or equal to $1$.

We still need to find the upper bound of $B_6$ and $B_7$.
\begin{align}\label{eq:B6}
B_6 
=  & ~  |\lambda_1 \cdot |v_1^\top \wh{v}_1|^p -\lambda_1|   \notag \\
=  & ~  \lambda_1 - \lambda_1 (1- \frac{1}{2} \| v_1 - \wh{v}_1 \|_2^2 )^p  \notag \\
 \leq  & ~ \lambda_1 p \frac{1}{2} \| v_1 - \wh{v}_1\|_2^2 \notag\\
 \leq & ~ p \epsilon^2 / (2 \lambda_1) \notag\\
 \leq & ~ \epsilon /12,
\end{align}
where the first step follows from the definition of $B_6$ (see Eq.~\eqref{eq:bound_lambda1-lambda1}), the second step follows from $v_1^\top \wh{v}_1  = 1 - \frac{1}{2} \| v_1 -\wh{v}_1 \|_2^2$, the third step comes from $\|v_1 -\wh{v}_1 \|_2^2 \ll 1$, the fourth step is because of $\|v_1 -\wh{v}_1 \|_2 \leq \epsilon /\lambda_1$, and the last step follows from $p \epsilon / (2\lambda_1) \leq 1/12$.

For $B_7$, we have
\begin{align}\label{eq:B7}
B_7 = & ~ \sum_{j=2}^k \lambda_j |v_j^\top \wh{v}_1|^p \notag\\
\leq & ~ \sum_{j=2}^k \lambda_j (\epsilon / (\sqrt{n}\lambda_j) )^p \notag\\
= & ~ \epsilon \sum_{j=2}^k (\epsilon / ( \lambda_j \sqrt{n}) )^{p-1} \notag\\
\leq & ~ \epsilon /4,
\end{align}
where the first step follows from the definition of $B_7$ (see Eq.~\eqref{eq:bound_lambda1-lambda1}), the second step follows from Part 3 of Definition~\ref{def:epsilon_close}, the third step follows from simple algebra, and the last step is due to $( \epsilon/ \lambda_k)^{p-1} \leq 1/ (4k)$. 

Let
\begin{align*}
    \epsilon < \frac{1}{4} k^{1/(p-1)} \lambda_k.
\end{align*}

Finally, combining everything together, we can get
\begin{align*}
|\wh{\lambda}_1 - \lambda_1| 
\leq & ~ B_5 + B_6 + B_7 \\
\leq & ~ \epsilon/12 + \epsilon/12 + \epsilon/4 \\
\leq & ~ \epsilon,
\end{align*}
where the first step follows from Eq.~\eqref{eq:bound_lambda1-lambda1}, the second step follows from combining Eq.~\eqref{eq:B5}, Eq.~\eqref{eq:B6}, and Eq.~\eqref{eq:B7}, and the last step follows from simple algebra.

{\bf Bounding $|\wh{v}_1^\top v_j|$.}

Let $j$ be an arbitrary element in $\{2,\cdots, k\}$. 

Let $t^*$ be the least integer satisfying 
\begin{align*}
    | v_1^\top u_{t^*}| \geq 1- \frac{1}{c_0^2 p^2 k^2},
\end{align*}
which implies 
\begin{align*}
    |v_j^\top u_{t^*}| \leq \frac{1}{c_0 pk}.
\end{align*}

By Part 3 of Theorem~\ref{thm:general_p_bound_EIu}, we have
\begin{align*}
 ~ |v_j^\top u_{t^*}| / |v_1^\top u_{t^*}| 
\leq & ~ 0.8 |v_j^\top u_{t^*-1}| / |v_1^\top u_{t^*-1}| \\
\leq & ~ \cdots \\
\leq & ~ 0.8^{t^*} \cdot |v_j^\top u_{0}| / |v_1^\top u_{0}| \\
\leq & ~ 0.8^{t^*} \cdot |v_j^\top u_{0}| / (1/\sqrt{n} ) \\
\leq & ~ 0.8^{t^*} \cdot 1 / (1/\sqrt{n} ),
\end{align*}
where the third step follows from recursively applying Part 3 of Theorem~\ref{thm:general_p_bound_EIu}, the fourth step follows from Lemma~\ref{lem:wang_lem_C1} and the last step follows from the fact that $|v_j^\top u_{0}|$ is at most 1.

Let
\begin{align*}
    t^* = \Omega(\log c_0 n).
\end{align*}

When $T> t^*$, we have
\begin{align*}
|v_j^\top u_{T}| / |v_1^\top u_{T}| \leq 0.8^{T-t^*}  |v_j^\top u_{t^*}| / |v_1^\top u_{t^*}| + 5 \cdot 18 \epsilon /(c_0 \lambda_1 \sqrt{n}).
\end{align*}

Let 
\begin{align*}
    T = \Omega(\log( n \lambda_1 /\epsilon))
\end{align*}
and $c_0 \geq 100$ to ensure
\begin{align*}
    |v_j^\top u_{T}| \leq \epsilon /(\lambda_1 \sqrt{n}).
\end{align*}

{\bf Inductive case.}

Let $i=r+1$.

Suppose the first $r$ cases holds.

To show the $r+1$ case also hold, we first consider the ``true'' noise, which is 
\begin{align*}
    \wt{E} = E+ \sum_{i=1}^r E_i + \ov{E}\in \mathbb{R}^{n^p}.
\end{align*}

We explain how to bound 
\begin{align*}
    \| \wh{v}_{r+1} - v_{r+1} \|_2,
\end{align*}
(for Definition~\ref{def:epsilon_close}, Part 2).  

Then, we show how to bound 
\begin{align*}
    | \wh{\lambda}_{r+1} - \lambda_{r+1} |
\end{align*}
as Part 1 of Definition~\ref{def:epsilon_close}. 

In the end, we show how to bound 
\begin{align*}
    |v_{r+1}^\top v_j|
\end{align*}
as Part 3 of Definition~\ref{def:epsilon_close}.

{\bf Bounding $\|\wh{v}_{r+1} -v_{r+1}\|_2$.}

Except for letting
\begin{align*}
T =\Omega(\log(n\lambda_{t+1}/\epsilon)),
\end{align*}
other parts of the proof are the same as the ones in the base case.

{\bf Bounding $|\wh{\lambda}_{r+1} -\lambda_{r+1}|$.}

Let $A ^*$ and $\wt{E}$ be 
\begin{align*}
    A ^* = \sum_{i=t+1}^k \lambda_i v_i^{\otimes p}
\end{align*}
and 
\begin{align*}
    \wt{E}= E + \ov{E} + \sum_{i=1}^t \wh{E}_i.
\end{align*}

Therefore, we have 
\begin{align*}
    |\wh{\lambda}_{t+1} -\lambda_{t+1}|
\end{align*}
satisfying
\begin{align*}
|\wh{\lambda}_{r+1} - \lambda_{r+1} |  
= & ~ | [A ^* +\wt{E}](\wh{v}_{r+1},\cdots, \wh{v}_{r+1}) - \lambda_{r+1} | \\
\leq & ~ |\wt{E}(\wh{v}_{r+1},\cdots, \wh{v}_{r+1})| + | A ^*(\wh{v}_{r+1},\cdots,\wh{v}_{r+1} ) - \lambda_{r+1}  |  \\
= & ~ |\wt{E}(\wh{v}_{r+1},\cdots, \wh{v}_{r+1} )| +  \left| \left[\sum_{i=r+1}^k \lambda_i v_i^{\otimes p}\right](\wh{v}_{r+1},\cdots,\wh{v}_{r+1} ) - \lambda_{r+1}  \right| \\
\leq & ~ \underbrace{ | \wt{E}(\wh{v}_{r+1},\cdots, \wh{v}_{r+1})| }_{B_8} + \underbrace{ |\lambda_{r+1} |v_{r+1}^\top \wh{v}_{r+1}|^p -\lambda_{r+1}| }_{B_9} + \underbrace{ \sum_{j=r+2}^k \lambda_j |v_j^\top \wh{v}_{r+1}|^p }_{B_{10}}.  
\end{align*}
where the first step follows from the definition of $\wh{\lambda}_{r+1}$, the second step follows from triangle inequality, the third step follows from $A ^* = \sum_{i=r+1}^k \lambda_i v_i^{\otimes p}$, and the last step follows from the triangle inequality.

We need to analyze $B_8$,
\begin{align*}
B_8 
= & ~  | \wt{E}(\wh{v}_{r+1},\cdots,\wh{v}_{r+1}) | \\
=  & ~ | E(\wh{v}_{r+1},\cdots,\wh{v}_{r+1}) | +| \ov{E} (\wh{v}_{r+1},\cdots, \wh{v}_{r+1}) | + | \sum_{i=1}^r \wh{E}_i(\wh{v}_{r+1},\cdots,\wh{v}_{r+1})  | \\
\leq & ~ \epsilon /( c_0 \sqrt{n})  + \epsilon / ( c_0\sqrt{n}) + 4 \epsilon/(c_0 \sqrt{n}) \\
\leq & ~ \epsilon/12, 
\end{align*}
where the first step follows from the definition of $B_8$, the second step follows from the triangle inequality, the third step follows from Eq.~\eqref{eq:bound_ovE}, the last step follows from $c_0 \geq 100, n\geq 1$.

$B_9$ and $B_{10}$ can be bounded in a similar way as the base case.

{\bf Bounding $|\wh{v}_{r+1}^\top v_j|$.}

Let $j$ be an arbitrary element in $\{ r+2, \cdots, k\}$. Then, the proof is the same as the base case.
\end{proof}

\begin{theorem}[Fast Tensor Power Method via Sketching, formal version of Theorem~\ref{thm:main_informal}]\label{thm:main_formal}

If the following conditions hold
\begin{itemize}
    \item Let $A  = A ^* +E \in \mathbb{R}^{n^p}$ be an arbitrary tensor satisfying $A ^* = \sum_{i=1}^k \lambda_i v_i^{\otimes p}$.
    \item Suppose that $\lambda_{1}$ is the greatest values in $\{\lambda_i\}_{i=1}^k$.
    \item Suppose that $\lambda_{k}$ is the smallest values in $\{\lambda_i\}_{i=1}^k$.
    \item The outputs obtained from the robust tensor power method are $\{\wh{\lambda}_i, \wh{v}_i\}_{i=1}^k$.
    \item $E$ satisfies that $\| E \| \leq \epsilon / (c_0 \sqrt{n} )$ (see definition of $\| E \|$ in Section~\ref{sec:prelim}).
    \item $T = \Omega( \log(\lambda_{1}  n/\epsilon) )$.
    \item $L = \Omega( k \log(k))$.
    \item $c_0\geq100$ and $c>0$
    \item For all $\epsilon$ satisfying $\epsilon \in (0, c \lambda_k / ( c_0  p^2 k n^{(p-2)/2})$.
\end{itemize}
 
  Then, our algorithm uses $\wt{O}(n^p)$ spaces, runs in $O(TL)$ iteration, and in each iteration it takes $\wt{O}(n^{p-1})$ time and then with probability at least $1-\delta$, there exists a permutation 
  \begin{align*}
      \pi:[k] \rightarrow [k],
  \end{align*}
  such that $\forall i\in[k]$,
  \begin{align*}
      |\lambda_i - \wh{\lambda}_{\pi(i)}| \leq \epsilon, \ \ \ \
  \|v_i - \wh{v}_{\pi(i)}\|_2 \leq \epsilon/ \lambda_i.
  \end{align*}
 \end{theorem}
\begin{proof}
It follows by combining Theorem~\ref{thm:general_p_without_sketch} and Lemma~\ref{lem:data_structure}.   
\end{proof}

%% file: sketch.tex
\section{Fast Sketching Data Structure}
\label{sec:fast_sketching_data_structure}

In this section, we introduce the fast sketching data structure

We state the definition of the data structure we will need.
\begin{definition}[Finding the top eigenvector and top-$k$ eigenvectors]
Given a collection of $n$ tensors $A_1, A_2, \cdots, A_n \in \R^{n^{p-1}}$, the goal is to design a structure that supports the following operations
\begin{itemize}
    \item \textsc{Init}  $(A_1, \cdots, A_n \in \R^{n^{p-1}})$. It takes $n$ tensors as inputs and create a data-structure.
    \item \textsc{Query}  $(u \in \R^n)$, the goal is to output a vector $v \in \R^n$ such that 
    \begin{align*} 
        v_i \approx \langle A_i, u^{\otimes (p-1)} \rangle, ~~~\forall i \in [n]
    \end{align*}
    \item \textsc{Query}$( \{ x_i \}_{i\in [k]} \in \R^n , \alpha \in \R^{n \times k}, u \in \R^n )$. the goal is to output a vector $v \in \R^n$ such that 
    \begin{align*}
    v_i \approx \langle A_i - \sum_{j=1}^k \alpha_{i,j} x_j^{\otimes (p-1)}  , u^{\otimes (p-1)} \rangle, ~~ \forall i \in [n]
    \end{align*}
\end{itemize}
\end{definition}

We state our data structure as follows:
\begin{lemma}[Data Structure]\label{lem:data_structure}

If the following conditions hold
\begin{itemize}
    \item Given $n$ matrices $A_1, A_2, \cdots A_n \in \R^{n^{p-1}}$ where $\| A_i \|_F \leq D_i$, $\forall i \in [n]$.
    \item Let $\| A \|_F \leq D$.
\end{itemize}

Then, there exists a randomized data structure with the following operations:

\begin{itemize}
    \item \textsc{Init}$(A_1, \cdots, A_n \in \R^{n ^{p-1}})$: It preprocesses $n$ tensors, in time $\wt O(\epsilon^{-2}n^{p}\log(1/\delta))$. 
    \item \textsc{Query}$(u \in \R^n)$. It takes a unit vector $u \in \R^n$ as input. The goal is to output a vector $v \in \R^n$ such that 
    \begin{align*}
   (1-\epsilon) \cdot \langle A_i, u^{\otimes (p-1)} \rangle - D_i \cdot \epsilon \leq  v_i \leq (1+\epsilon) \cdot \langle A_i, u^{\otimes (p-1)} \rangle + D_i \cdot \epsilon, \forall i \in [n].
    \end{align*}
    This can be done in time $\wt O(\epsilon^{-2}n^{(p-1)}\log(1/\delta))$.
    \item \textsc{QueryValue}$(u \in \R^n)$. The goal is to output a number $v \in \R$ such that
    \begin{align*}
        (1-\epsilon) \langle A, u^{\otimes p} \rangle - D \cdot \epsilon \leq v \leq (1+\epsilon) \langle A, u^{\otimes p} \rangle + D \cdot \epsilon .
    \end{align*}
    This can be done in time $\wt O(\epsilon^{-2}n^{(p-1)}\log(1/\delta))$.
    \item \textsc{QueryRes}$( \{ x_j \}_{j\in [k]} \in \R^n , \alpha \in \R^{n \times k}, u \in \R^n )$. The goal is to output a vector $v \in \R^n$ such that 
    \begin{align*}
    & ~ (1-\epsilon) \cdot \langle A_i - \sum_{j=1}^k \alpha_{i,j} x_j^{\otimes (p-1)} , u^{\otimes (p-1)} \rangle - D_i \cdot \epsilon \\
    & ~ \leq v_i \leq (1+\epsilon) \cdot \langle A_i - \sum_{j=1}^k \alpha_{i,j} x_j^{\otimes (p-1)}  , u^{\otimes (p-1)} \rangle + D_i \cdot \epsilon, \forall i \in [n]
    \end{align*}
    This can be done in time $\wt O(\epsilon^{-2}n^{(p-1)}\log(1/\delta)+n^{2} k)$.
\end{itemize}
All the queries are robust to adversary type queries.
\end{lemma}
\begin{proof}
The correctness of \textsc{Init} and $\textsc{Query}$ directly follows from \cite{cn22}.

For the \textsc{QueryResidual}, the running time only need to pay an extra term is computing
\begin{align*}
\langle \sum_{j=1}^k \alpha_{i,j} x_j^{\otimes (p-1)} , u^{\otimes (p-1)} \rangle 
\end{align*}
which is sufficient just to compute
\begin{align*}
\sum_{j=1}^k \alpha_{i,j} \langle x_j, u \rangle^{p-1}.
\end{align*}
The above step takes $O(kn)$ time. Since there are $n$ different indices $i$. So overall extra time is $O(n^2 k)$.
\end{proof}

%% file: main.bbl
\newcommand{\etalchar}[1]{$^{#1}$}
\begin{thebibliography}{SVBDL13}

\bibitem[AGH{\etalchar{+}}12]{agh+12}
Anima Anandkumar, Rong Ge, Daniel Hsu, Sham~M. Kakade, and Matus Telgarsky.
\newblock Tensor decompositions for learning latent variable models.
\newblock 2012.

\bibitem[AGH{\etalchar{+}}14]{aghkt14}
Animashree Anandkumar, Rong Ge, Daniel~J. Hsu, Sham~M. Kakade, and Matus
  Telgarsky.
\newblock Tensor decompositions for learning latent variable models.
\newblock In {\em Journal of Machine Learning Research}, volume 15(1), pages
  2773--2832. \url{https://arxiv.org/pdf/1210.7559}, 2014.

\bibitem[AGJ17]{agj17}
Animashree Anandkumar, Rong Ge, and Majid Janzamin.
\newblock Analyzing tensor power method dynamics in overcomplete regime.
\newblock {\em Journal of Machine Learning Research}, 18(22):1--40, 2017.

\bibitem[ALA16]{ala16}
Kamyar Azizzadenesheli, Alessandro Lazaric, and Animashree Anandkumar.
\newblock Reinforcement learning of {POMDP}s using spectral methods.
\newblock In {\em 29th Annual Conference on Learning Theory (COLT)}, pages
  193--256. \url{https://arxiv.org/pdf/1602.07764}, 2016.

\bibitem[AMR09]{amr09}
Elizabeth~S. Allman, Catherine Matias, and John~A. Rhodes.
\newblock Identifiability of parameters in latent structure models with many
  observed variables.
\newblock {\em The Annals of Statistics}, 37(6A), dec 2009.

\bibitem[APRS09]{aprs11}
Elizabeth~S. Allman, Sonja Petrović, John~A. Rhodes, and Seth Sullivant.
\newblock Identifiability of 2-tree mixtures for group-based models, 2009.

\bibitem[BKS21]{bks21}
Vineet Bhatt, Sunil Kumar, and Seema Saini.
\newblock Tucker decomposition and applications.
\newblock {\em Materials Today: Proceedings}, 46:10787--10792, 2021.
\newblock International Conference on Technological Advancements in Materials
  Science and Manufacturing.

\bibitem[BS15]{bs15}
Srinadh Bhojanapalli and Sujay Sanghavi.
\newblock A new sampling technique for tensors.
\newblock In {\em arXiv preprint}. \url{https://arxiv.org/pdf/1502.05023},
  2015.

\bibitem[BV18]{bv18}
Paul Breiding and Nick Vannieuwenhoven.
\newblock A riemannian trust region method for the canonical tensor rank
  approximation problem.
\newblock {\em SIAM Journal on Optimization}, 28(3):2435--2465, 2018.

\bibitem[CC70]{cc70}
J~Douglas Carroll and Jih-Jie Chang.
\newblock Anaylsis of individual differences in multidimensional scaling via an
  n-way generalization of eckart-young decomposition.
\newblock {\em Psychometrika}, 35(3):283--319, 1970.

\bibitem[CLS19]{cls19}
Michael~B Cohen, Yin~Tat Lee, and Zhao Song.
\newblock Solving linear programs in the current matrix multiplication time.
\newblock In {\em Proceedings of the 51st Annual ACM Symposium on Theory of
  Computing (STOC)}. \url{https://arxiv.org/pdf/1810.07896.pdf}, 2019.

\bibitem[CN22]{cn22}
Yeshwanth Cherapanamjeri and Jelani Nelson.
\newblock Uniform approximations for randomized hadamard transforms with
  applications.
\newblock {\em arXiv preprint arXiv:2203.01599}, 2022.

\bibitem[CPZ08]{cpz08}
Kung-Ching Chang, Kelly Pearson, and Tan Zhang.
\newblock Perron-frobenius theorem for nonnegative tensors.
\newblock {\em Communications in Mathematical Sciences}, 6(2):507--520, 2008.

\bibitem[CV14]{cv14}
Joon~Hee Choi and S.~Vishwanathan.
\newblock Dfacto: Distributed factorization of tensors.
\newblock In {\em NIPS}, pages 1296--1304, 2014.

\bibitem[DGS23]{dgs23}
Yichuan Deng, Yeqi Gao, and Zhao Song.
\newblock Solving tensor low cycle rank approximation.
\newblock {\em arXiv preprint arXiv:2304.06594}, 2023.

\bibitem[DJS{\etalchar{+}}22]{djs+22}
Yichuan Deng, Wenyu Jin, Zhao Song, Xiaorui Sun, and Omri Weinstein.
\newblock Dynamic kernel sparsifiers.
\newblock {\em arXiv preprint arXiv:2211.14825}, 2022.

\bibitem[DKN22]{dkn22}
Dipak Dulal, Ramin~Goudarzi Karim, and Carmeliza Navasca.
\newblock Covid-19 analysis using tensor methods, 2022.

\bibitem[DLS23]{dls23}
Yichuan Deng, Zhihang Li, and Zhao Song.
\newblock An improved sample complexity for rank-1 matrix sensing.
\newblock {\em arXiv preprint arXiv:2303.06895}, 2023.

\bibitem[DLY21]{dly21}
Sally Dong, Yin~Tat Lee, and Guanghao Ye.
\newblock A nearly-linear time algorithm for linear programs with small
  treewidth: a multiscale representation of robust central path.
\newblock In {\em Proceedings of the 53rd Annual ACM SIGACT Symposium on Theory
  of Computing}, pages 1784--1797, 2021.

\bibitem[DMS23]{dms23}
Yichuan Deng, Sridhar Mahadevan, and Zhao Song.
\newblock Randomized and deterministic attention sparsification algorithms for
  over-parameterized feature dimension.
\newblock {\em arXiv preprint arXiv:2304.04397}, 2023.

\bibitem[DSWZ22]{dswz22}
Yichuan Deng, Zhao Song, Omri Weinstein, and Ruizhe Zhang.
\newblock Fast distance oracles for any symmetric norm.
\newblock {\em arXiv preprint arXiv:2205.14816}, 2022.

\bibitem[GQSW22]{gqls23}
Yeqi Gao, Lianke Qin, Zhao Song, and Yitan Wang.
\newblock A sublinear adversarial training algorithm.
\newblock {\em arXiv preprint arXiv:2208.05395}, 2022.

\bibitem[GS22]{gs22}
Yuzhou Gu and Zhao Song.
\newblock A faster small treewidth sdp solver.
\newblock {\em arXiv preprint arXiv:2211.06033}, 2022.

\bibitem[GSY23]{gsy23}
Yeqi Gao, Zhao Song, and Junze Yin.
\newblock An iterative algorithm for rescaled hyperbolic functions regression.
\newblock {\em arXiv preprint arXiv:2305.00660}, 2023.

\bibitem[GSYZ23]{gsyz23}
Yuzhou Gu, Zhao Song, Junze Yin, and Lichen Zhang.
\newblock Low rank matrix completion via robust alternating minimization in
  nearly linear time.
\newblock {\em arXiv preprint arXiv:2302.11068}, 2023.

\bibitem[Har70]{h70}
Richard~A Harshman.
\newblock Foundations of the parafac procedure: Models and conditions for an"
  explanatory" multimodal factor analysis.
\newblock 1970.

\bibitem[HCL22]{hcl22}
Qiang Heng, Eric~C. Chi, and Yufeng Liu.
\newblock Tucker-$\operatorname{L_2E}$: Robust low-rank tensor decomposition
  with the $\operatorname{L_2}$ criterion, 2022.

\bibitem[HH82]{hh82}
Chikio Hayashi and Fumi Hayashi.
\newblock A new algorithm to solve parafac-model.
\newblock {\em Behaviormetrika}, 9(11):49--60, 1982.

\bibitem[HL13]{hl13}
Christopher~J Hillar and Lek-Heng Lim.
\newblock Most tensor problems are np-hard.
\newblock In {\em Journal of the ACM (JACM)}, volume 60(6), page~45.
  \url{https://arxiv.org/pdf/0911.1393}, 2013.

\bibitem[HNH{\etalchar{+}}13]{hnh+13}
Furong Huang, Niranjan~U. N, Mohammad~Umar Hakeem, Prateek Verma, and
  Animashree Anandkumar.
\newblock Fast detection of overlapping communities via online tensor methods
  on gpus.
\newblock {\em CoRR}, abs/1309.0787, 2013.

\bibitem[JSA15]{jsa15}
Majid Janzamin, Hanie Sedghi, and Anima Anandkumar.
\newblock Beating the perils of non-convexity: Guaranteed training of neural
  networks using tensor methods.
\newblock In {\em arXiv preprint}. \url{https://arxiv.org/pdf/1506.08473},
  2015.

\bibitem[JSWZ21]{jswz21}
Shunhua Jiang, Zhao Song, Omri Weinstein, and Hengjie Zhang.
\newblock Faster dynamic matrix inverse for faster lps.
\newblock In {\em STOC}, 2021.

\bibitem[KB09]{kb09}
Tamara~G. Kolda and Brett~W. Bader.
\newblock Tensor decompositions and applications.
\newblock {\em {SIAM} Review}, 51(3):455--500, 2009.

\bibitem[KC16]{kc16}
Mijung Kim and K.~Selçuk Candan.
\newblock Decomposition-by-normalization (dbn): leveraging approximate
  functional dependencies for efficient cp and tucker decompositions.
\newblock 30(1):1--46, 2016.

\bibitem[KM11]{km11}
Tamara~G Kolda and Jackson~R Mayo.
\newblock Shifted power method for computing tensor eigenpairs.
\newblock {\em SIAM Journal on Matrix Analysis and Applications},
  32(4):1095--1124, 2011.

\bibitem[KMM22]{kmm22}
Rima Khouja, Pierre-Alexandre Mattei, and Bernard Mourrain.
\newblock Tensor decomposition for learning gaussian mixtures from moments.
\newblock {\em Journal of Symbolic Computation}, 113:193--210, 2022.

\bibitem[KPHF12]{kphf12}
U.~Kang, Evangelos~E. Papalexakis, Abhay Harpale, and Christos Faloutsos.
\newblock Gigatensor: scaling tensor analysis up by 100 times - algorithms and
  discoveries.
\newblock In {\em KDD}, pages 316--324, 2012.

\bibitem[Kru77]{k77}
Joseph~B. Kruskal.
\newblock Three-way arrays: rank and uniqueness of trilinear decompositions,
  with application to arithmetic complexity and statistics.
\newblock {\em Linear Algebra and its Applications}, 18(2):95--138, 1977.

\bibitem[Lim05]{l05}
Lek-Heng Lim.
\newblock Singular values and eigenvalues of tensors: a variational approach.
\newblock In {\em 1st IEEE International Workshop on Computational Advances in
  Multi-Sensor Adaptive Processing, 2005.}, pages 129--132. IEEE, 2005.

\bibitem[LSZ19]{lsz19}
Yin~Tat Lee, Zhao Song, and Qiuyi Zhang.
\newblock Solving empirical risk minimization in the current matrix
  multiplication time.
\newblock In {\em COLT}. \url{https://arxiv.org/pdf/1905.04447.pdf}, 2019.

\bibitem[LSZ23]{lsz23}
Zhihang Li, Zhao Song, and Tianyi Zhou.
\newblock Solving regularized exp, cosh and sinh regression problems.
\newblock {\em arXiv preprint arXiv:2303.15725}, 2023.

\bibitem[MWZ22]{mwz22}
Arvind~V Mahankali, David~P Woodruff, and Ziyu Zhang.
\newblock Near-linear time and fixed-parameter tractable algorithms for tensor
  decompositions.
\newblock {\em arXiv preprint arXiv:2207.07417}, 2022.

\bibitem[NQZ10]{nqz10}
Michael Ng, Liqun Qi, and Guanglu Zhou.
\newblock Finding the largest eigenvalue of a nonnegative tensor.
\newblock {\em SIAM Journal on Matrix Analysis and Applications},
  31(3):1090--1099, 2010.

\bibitem[OT09]{ot09}
Ivan~V Oseledets and Eugene~E Tyrtyshnikov.
\newblock Breaking the curse of dimensionality, or how to use svd in many
  dimensions.
\newblock {\em SIAM journal on scientific computing}, 31(5):3744--3759, 2009.

\bibitem[Paa99]{p99}
Pentti Paatero.
\newblock The multilinear engine—a table-driven, least squares program for
  solving multilinear problems, including the n-way parallel factor analysis
  model.
\newblock {\em Journal of Computational and Graphical Statistics},
  8(4):854--888, 1999.

\bibitem[PC08]{pc08}
Anh Phan and Andrzej Cichocki.
\newblock Fast and efficient algorithms for nonnegative tucker decomposition.
\newblock {\em Advances in Neural Networks-ISNN 2008}, pages 772--782, 2008.

\bibitem[PTC13]{ptc13}
Anh-Huy Phan, Petr Tichavsky, and Andrzej Cichocki.
\newblock Low complexity damped gauss--newton algorithms for candecomp/parafac.
\newblock {\em SIAM Journal on Matrix Analysis and Applications},
  34(1):126--147, 2013.

\bibitem[Qi05]{q05}
Liqun Qi.
\newblock Eigenvalues of a real supersymmetric tensor.
\newblock {\em Journal of Symbolic Computation}, 40(6):1302--1324, 2005.

\bibitem[Qi07]{q07}
Liqun Qi.
\newblock Eigenvalues and invariants of tensors.
\newblock {\em Journal of Mathematical Analysis and Applications},
  325(2):1363--1377, 2007.

\bibitem[QJS{\etalchar{+}}22]{qjs+22}
Lianke Qin, Rajesh Jayaram, Elaine Shi, Zhao Song, Danyang Zhuo, and Shumo Chu.
\newblock Adore: Differentially oblivious relational database operators.
\newblock {\em VLDB}, 2022.

\bibitem[QRS{\etalchar{+}}22]{qrs+22}
Lianke Qin, Aravind Reddy, Zhao Song, Zhaozhuo Xu, and Danyang Zhuo.
\newblock Adaptive and dynamic multi-resolution hashing for pairwise
  summations.
\newblock {\em arXiv preprint arXiv:2212.11408}, 2022.

\bibitem[QSW07]{qsw07}
Liqun Qi, Wenyu Sun, and Yiju Wang.
\newblock Numerical multilinear algebra and its applications.
\newblock {\em Frontiers of Mathematics in China}, 2:501--526, 2007.

\bibitem[QSW23]{qsw23}
Lianke Qin, Zhao Song, and Yitan Wang.
\newblock Fast submodular function maximization.
\newblock {\em arXiv preprint arXiv:2305.08367}, 2023.

\bibitem[QSZ23]{qsz23}
Lianke Qin, Zhao Song, and Ruizhe Zhang.
\newblock A general algorithm for solving rank-one matrix sensing.
\newblock {\em arXiv preprint arXiv:2303.12298}, 2023.

\bibitem[QSZZ23]{qszz23}
Lianke Qin, Zhao Song, Lichen Zhang, and Danyang Zhuo.
\newblock An online and unified algorithm for projection matrix vector
  multiplication with application to empirical risk minimization.
\newblock In {\em International Conference on Artificial Intelligence and
  Statistics}, pages 101--156. PMLR, 2023.

\bibitem[RS12]{rs12}
John~A Rhodes and Seth Sullivant.
\newblock Identifiability of large phylogenetic mixture models.
\newblock {\em Bulletin of mathematical biology}, 74:212--231, 2012.

\bibitem[RSG17]{rsg17}
Stephan Rabanser, Oleksandr Shchur, and Stephan Günnemann.
\newblock Introduction to tensor decompositions and their applications in
  machine learning, 2017.

\bibitem[RSZ22]{rsz22}
Aravind Reddy, Zhao Song, and Lichen Zhang.
\newblock Dynamic tensor product regression.
\newblock {\em arXiv preprint arXiv:2210.03961}, 2022.

\bibitem[SH05]{sh05}
Amnon Shashua and Tamir Hazan.
\newblock Non-negative tensor factorization with applications to statistics and
  computer vision.
\newblock In {\em Proceedings of the 22nd international conference on Machine
  learning(ICML)}, pages 792--799. ACM, 2005.

\bibitem[SL10]{sl10}
Berkant Savas and Lek-Heng Lim.
\newblock Quasi-newton methods on grassmannians and multilinear approximations
  of tensors.
\newblock {\em SIAM Journal on Scientific Computing}, 32(6):3352--3393, 2010.

\bibitem[SVBDL13]{sbl13}
Laurent Sorber, Marc Van~Barel, and Lieven De~Lathauwer.
\newblock Optimization-based algorithms for tensor decompositions: Canonical
  polyadic decomposition, decomposition in rank-(l\_r,l\_r,1) terms, and a new
  generalization.
\newblock {\em SIAM Journal on Optimization}, 23(2):695--720, 2013.

\bibitem[SWZ16]{swz16}
Zhao Song, David~P. Woodruff, and Huan Zhang.
\newblock Sublinear time orthogonal tensor decomposition.
\newblock In {\em Advances in Neural Information Processing Systems 29: Annual
  Conference on Neural Information Processing Systems (NIPS) 2016, December
  5-10, 2016, Barcelona, Spain}, pages 793--801, 2016.

\bibitem[SWZ18]{swz18}
Zhao Song, David~P Woodruff, and Peilin Zhong.
\newblock Towards a zero-one law for entrywise low rank approximation.
\newblock {\em arXiv preprint arXiv:1811.01442}, 2018.

\bibitem[SWZ19]{swz19}
Zhao Song, David~P Woodruff, and Peilin Zhong.
\newblock Relative error tensor low rank approximation.
\newblock In {\em SODA}. arXiv preprint arXiv:1704.08246, 2019.

\bibitem[SXZ22]{sxz22}
Zhao Song, Zhaozhuo Xu, and Lichen Zhang.
\newblock Speeding up sparsification using inner product search data
  structures.
\newblock {\em arXiv preprint arXiv:2204.03209}, 2022.

\bibitem[SYYZ22]{syyz22_lichen}
Zhao Song, Xin Yang, Yuanyuan Yang, and Lichen Zhang.
\newblock Sketching meets differential privacy: Fast algorithm for dynamic
  kronecker projection maintenance.
\newblock {\em arXiv preprint arXiv:2210.11542}, 2022.

\bibitem[TB06]{tb06}
Giorgio Tomasi and Rasmus Bro.
\newblock A comparison of algorithms for fitting the parafac model.
\newblock {\em Computational Statistics \& Data Analysis}, 50(7):1700--1734,
  2006.

\bibitem[Tso10]{t10}
Charalampos~E. Tsourakakis.
\newblock {MACH:} fast randomized tensor decompositions.
\newblock In {\em SDM}, pages 689--700, 2010.

\bibitem[TT21]{tt21}
Y-H. Taguchi and Turki Turki.
\newblock Application of tensor decomposition to gene expression of infection
  of mouse hepatitis virus can identify critical human genes and efffective
  drugs for sars-cov-2 infection.
\newblock {\em IEEE Journal of Selected Topics in Signal Processing},
  15(3):746--758, 2021.

\bibitem[WA16]{wa16}
Yining Wang and Animashree Anandkumar.
\newblock Online and differentially-private tensor decomposition.
\newblock In {\em Advances in Neural Information Processing Systems 29: Annual
  Conference on Neural Information Processing Systems (NIPS) 2016, December
  5-10, 2016, Barcelona, Spain}. \url{https://arxiv.org/pdf/1606.06237}, 2016.

\bibitem[WLSH14]{wlsh14}
Chi Wang, Xueqing Liu, Yanglei Song, and Jiawei Han.
\newblock Scalable moment-based inference for latent dirichlet allocation.
\newblock In {\em ECML-PKDD}, pages 290--305, 2014.

\bibitem[WQZ09]{wqz09}
Yiju Wang, Liqun Qi, and Xinzhen Zhang.
\newblock A practical method for computing the largest m-eigenvalue of a
  fourth-order partially symmetric tensor.
\newblock {\em Numerical Linear Algebra with Applications}, 16(7):589--601,
  2009.

\bibitem[WTSA15]{wtsa15}
Yining Wang, Hsiao-Yu Tung, Alexander~J Smola, and Anima Anandkumar.
\newblock Fast and guaranteed tensor decomposition via sketching.
\newblock In {\em Advances in Neural Information Processing Systems (NIPS)},
  pages 991--999. \url{https://arxiv.org/pdf/1506.04448}, 2015.

\bibitem[Ye20]{y20}
Guanghao Ye.
\newblock Fast algorithm for solving structured convex programs.
\newblock {\em The University of Washington, Undergraduate Thesis}, 2020.

\bibitem[ZCZX15]{zczx15}
Guoxu Zhou, Andrzej Cichocki, Qibin Zhao, and Shengli Xie.
\newblock Efficient nonnegative tucker decompositions: Algorithms and
  uniqueness.
\newblock {\em {IEEE} Transactions on Image Processing}, 24(12):4990--5003, dec
  2015.

\bibitem[ZD13]{mc13}
Miao Zhang and Chris Ding.
\newblock Robust tucker tensor decomposition for effective image
  representation.
\newblock In {\em 2013 IEEE International Conference on Computer Vision}, pages
  2448--2455, 2013.

\end{thebibliography}
